\documentclass[journal,twoside]{IEEEtran}

\usepackage{amsmath,amsfonts}
\usepackage{algorithmic}
\usepackage{algorithm}
\usepackage{array}
\usepackage[caption=false,font=footnotesize,labelfont=rm,textfont=rm]{subfig}
\usepackage{textcomp}
\usepackage{stfloats}
\usepackage{url}
\usepackage{verbatim}
\usepackage{graphicx}
\usepackage{cite}

\usepackage[breaklinks,colorlinks, linkcolor=red, citecolor=blue, urlcolor=black]{hyperref}

\usepackage[utf8]{inputenc} 
\usepackage[T1]{fontenc}    
\usepackage{hyperref}       
\usepackage[hyphenbreaks]{breakurl}
\usepackage{url}            
\usepackage{booktabs}       
\usepackage{amsfonts}       
\usepackage{nicefrac}       
\usepackage{microtype}      
\usepackage{xcolor}         
\usepackage{multirow}
\usepackage{graphicx}
\usepackage{amsmath}
\usepackage{amssymb}
\usepackage{utfsym}
\usepackage{pifont}
\usepackage{bm}
\usepackage{float}
\usepackage{subfig}
\usepackage{wrapfig}
\newcommand{\ie}{{\emph{i.e.}}}
\newcommand{\eg}{{\emph{e.g.}}}
\newcommand{\etal}{{\emph{et al.}}}

\newcommand{\cmark}{\ding{51}}%
\newcommand{\xmark}{\ding{55}}
\newenvironment{tightcenter}{%
  \setlength\topsep{3pt}
  \setlength\parskip{3pt}
  \begin{center}
  \begin{minipage}{.4\textwidth}
}{%
  \end{minipage}
  \end{center}
}

\DeclareMathOperator*{\argmin}{arg\,min}

\usepackage{xcolor}
\usepackage{colortbl}

\begin{document}

\title{Semantic Contextualization of Face Forgery: A New Definition, Dataset, and Detection Method}

\author{Mian~Zou,
        Baosheng~Yu,
        Yibing~Zhan,~\IEEEmembership{Member,~IEEE,}
        Siwei~Lyu,~\IEEEmembership{Fellow,~IEEE,}
        and~Kede~Ma,~\IEEEmembership{Senior Member,~IEEE}
\thanks{Mian Zou and Kede Ma are with the Department of Computer Science, City University of Hong Kong, Kowloon, Hong Kong (e-mail: mianzou2-c@my.cityu.edu.hk; kede.ma@cityu.edu.hk).}
\thanks{Baosheng Yu is with the Lee Kong Chian School of Medicine, Nanyang Technological University, Singapore (e-mail: baosheng.yu@ntu.edu.sg).}
\thanks{Yibing Zhan is with the JD Explore Academy, Beijing, China (e-mail: zhanyibing@jd.com).}
\thanks{Siwei Lyu is with the Department of Computer Science and Engineering, University at Buffalo, State University of New York, Buffalo, NY USA (e-mail: siweilyu@buffalo.edu).}
\thanks{\emph{Corresponding author: Kede~Ma}.}
}

\markboth{IEEE Transactions on Information Forensics and Security}%
{Zou \MakeLowercase{\textit{et al.}}: Semantics-Oriented Face Forgery Detection}


\maketitle

\begin{abstract}
In recent years, deep learning has greatly streamlined the process of manipulating photographic face images. Aware of the potential dangers, researchers have developed various tools to spot these counterfeits. Yet, none asks the fundamental question: \textit{What digital manipulations make a real photographic face image fake, while others do not?}
In this paper, we put face forgery in a semantic context and define that \textit{computational methods that alter semantic face attributes to exceed human discrimination thresholds are sources of face forgery}. Following our definition, we construct a large face forgery image dataset, where each image is associated with a set of labels organized in a hierarchical graph. Our dataset enables two new testing protocols to probe the generalizability of face forgery detectors. Moreover, we propose a semantics-oriented face forgery detection method that captures label relations and prioritizes the primary task (\ie, real or fake face detection). We show that the proposed dataset successfully exposes the weaknesses of current detectors as the test set and consistently improves their generalizability as the training set.  Additionally, we demonstrate the superiority of our semantics-oriented method over traditional binary and multi-class classification-based detectors.
\end{abstract}

\begin{IEEEkeywords}
Face forgery detection, face semantics, datasets.
\end{IEEEkeywords}

\section{Introduction}~\label{sec:intro}
\IEEEPARstart{T}{he} recent strides in deep learning~\cite{faceswap,song2019generative,karras2020analyzing} have significantly facilitated face forgery~\cite{farid2016photo_forensics}. Alongside its entertaining applications, face forgery sparks widespread public anxieties due to reported misuses. Instances include the generation of nonconsensual pornography, biometric fraud, service disruption, and political manipulation.

While several face forgery detection tools have been created, their generalizability to novel face manipulations remains limited. Early detectors rely primarily on three types of knowledge. The first is knowledge about statistical regularities of real face images~\cite{popescu2005exposing, popescu2004statistical}.
The second is knowledge about the digital photography pipeline, in which various visual artifacts may be characterized to expose forgeries~\cite{popescu2005exposing2, johnson2006exposing, lyu2010estimating, lin2005detecting, lyu2014exposing}. The
third is knowledge about the 3D world we are living in, particularly the physical laws that govern the interactions of light, optics, and objects~\cite{johnson2007exposing, kee2014exposing, o2012exposing, conotter2011exposing}. However, designing computational structures manually to exploit domain knowledge is a highly challenging task. Consequently, existing knowledge-driven methods are often tailored to specific forgery scenarios, resulting in limited generalizability.
With the advent of deep learning, data-driven detectors~\cite{Li2018in, afchar2018mesonet, zhou2017two, Nguyen2019multi, li2020face, zhao2021multi, dong2022protecting, Dong_2023_CVPR} have come to the forefront, whose effectiveness is attributed to the quality of training data and the formulation of face forgery detection.

Typically, face forgery detection is formulated as a standard binary classification problem. A natural extension is multi-class (\ie, $C$-way) classification~\cite{he2021forgerynet, Sun_2023_ICCV}, in which $C-1$ classes are designated for $C-1$ different types of fake manipulations, while the remaining class is dedicated to encompassing all real manipulations. 
Despite the demonstrated success, a fundamental question in this field has been treated superficially:
\begin{tightcenter}
\textit{What digital manipulations make a real photographic face image fake, while others do not?}
\end{tightcenter}
If we are unable to draw a boundary between real and fake manipulations, it is not possible to discuss the generalizability of face forgery detectors. Previous generalization tests involve training detectors on some ``deemed fake'' manipulations (\eg, Deepfakes~\cite{faceswap}) and measuring performance on another set of ``deemed fake'' manipulations
(\eg, Face2Face~\cite{thies2016face2face}). Clearly, this setting does not accurately reflect the complexities of real-world scenarios.

\begin{figure*}[t]
  \centering
   \subfloat[Original face]{\includegraphics[width=0.14\linewidth]{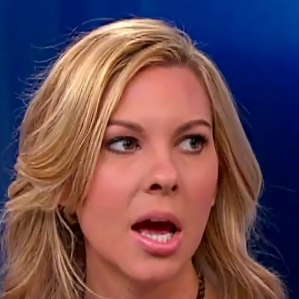}}\hskip.1em
   \subfloat[$d_\mathrm{age}=-0.05$]{\includegraphics[width=0.14\linewidth]{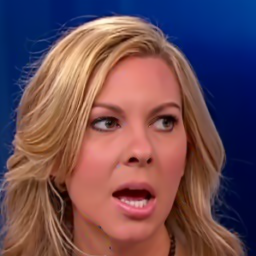}}\hskip.1em
   \subfloat[$d_\mathrm{age}=-0.10$]{\includegraphics[width=0.14\linewidth]{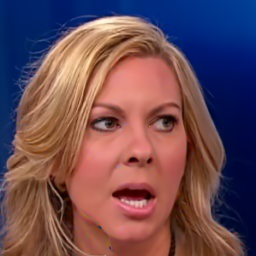}}\hskip.1em
   \subfloat[$d_\mathrm{age}=-0.15$]{\includegraphics[width=0.14\linewidth]{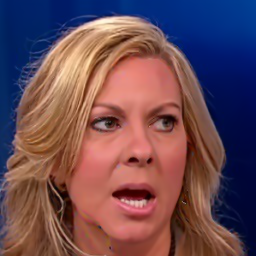}}\hskip.1em
   \subfloat[$d_\mathrm{age}=-0.20$]{\includegraphics[width=0.14\linewidth]{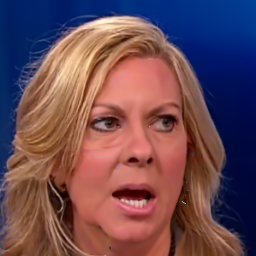}}\hskip.1em
   \subfloat[$d_\mathrm{age}=-0.25$]{\includegraphics[width=0.14\linewidth]{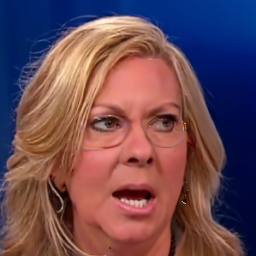}}\hskip.1em
   \subfloat[$d_\mathrm{age}=-0.30$]{\includegraphics[width=0.14\linewidth]{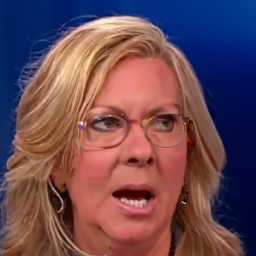}}
   \caption{Illustration of diffusion autoencoders~\cite{Preechakul_2022_CVPR} on age manipulation. By varying the age parameter $d_\mathrm{age}$, which controls the movement of the image latent along the age direction, we create a set of age-manipulated images, only a subset of which are considered fake according to our definition (\eg, those with $d_\mathrm{age} \le -0.20$). A more negative number indicates an older age.}
  \label{fig: degree_control_fakeness}

\end{figure*}

In this paper, we rethink face forgery from both conceptual and computational perspectives. We first define face forgery in a semantic context:
\begin{tightcenter}
\textit{Computational methods that alter semantic face attributes to exceed human discrimination thresholds are sources of face forgery.}
\end{tightcenter}
Here, the description ``to exceed human discrimination thresholds'' means that the alteration of semantic face attributes relative to the original real photographic image is discernable to the human eye.
Fig.~\ref{fig: degree_control_fakeness} presents an example of diffusion autoencoders~\cite{Preechakul_2022_CVPR} on age manipulation. By moving the image latent along the age direction\footnote{Such a direction can be identified as the weight vector of a linear classifier trained on latent codes of positive and negative images of the age attribute~\cite{Preechakul_2022_CVPR}.}, \ie, tuning the age parameter $d_\mathrm{age}$,
we can generate a sequence of manipulated images, of which some are easily distinguishable from the original image\footnote{The original real photographic face image is chosen from the FaceForensics++ dataset~\cite{rossler2019faceforensics} for illustration purposes only.} (\eg,  $d_\mathrm{age}\le -0.20$). 
In contrast, images with the age parameter greater than $-0.10$ are innocuous in real-world applications because they retain nearly all semantic face attributes.

\begin{figure}[t]
  \centering
   \subfloat[]{\includegraphics[width=0.9\linewidth]{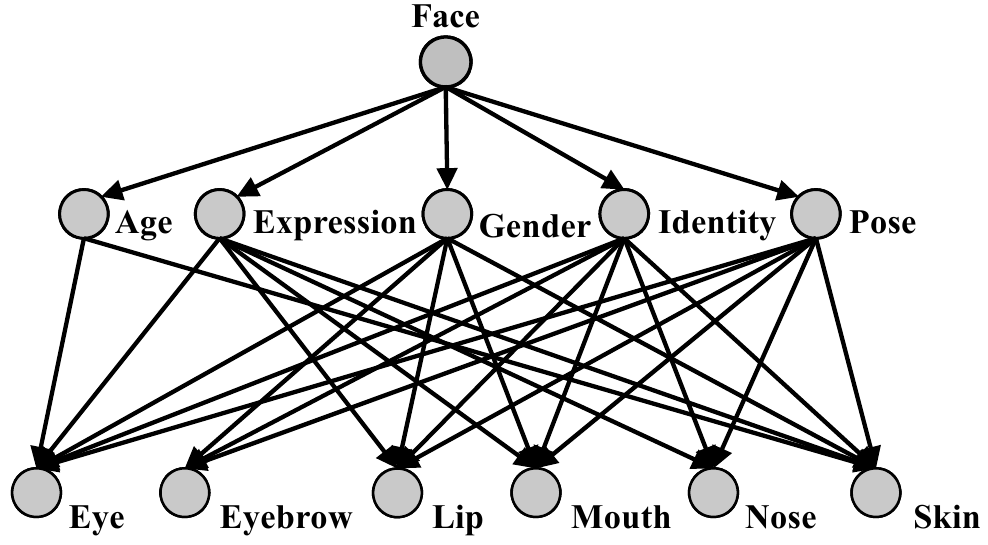}} \\
   \subfloat[]{\includegraphics[width=0.9\linewidth]{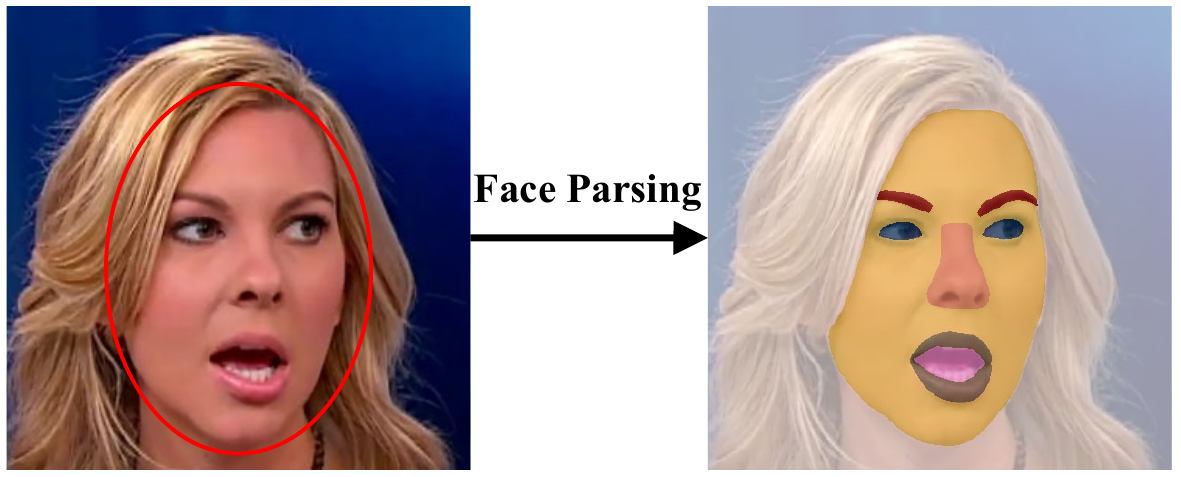}}
   \caption{\textbf{(a)} Hierarchical graph for label relation encoding in FFSC. We partition the root node, denoted as \texttt{face}, into five global face attribute nodes, \texttt{age}, \texttt{expression}, \texttt{gender}, \texttt{identity}, and \texttt{pose}, each of which is further connected to a set of leaf nodes, representing local face regions. \textbf{(b)} Parsing of the face in Fig.~\ref{fig: degree_control_fakeness}(a) into non-overlapping  local face regions.
   }
  \label{fig_DAG}
  
\end{figure}

Guided by our definition, we create a new Face Forgery in the Semantic Context (FFSC) dataset, in which each image is associated with a set of semantic labels organized in a hierarchical acyclic graph, as shown in Fig.~\ref{fig_DAG}. We contextualize twelve popular face manipulations~\cite{Preechakul_2022_CVPR, Pehlivan_2023_CVPR, First_Order_Motion_nips, viazovetskyi2020stylegan2, chen2020simswap, nirkin2019fsgan, li2020face, zhao2022thin, Patashnik_2021_ICCV, Gao_2021_CVPR, zeng2022fnevr, wang2021high} into five global face attributes (\ie, \texttt{age}, \texttt{expression}, \texttt{gender}, \texttt{identity}, and \texttt{pose}). Notably, a single manipulation method can modify multiple attributes, while multiple different methods can modify the same face attribute. In FFSC, the manipulation degree that exceeds human discrimination thresholds and the connectivity to local face regions (\ie, \texttt{eye}, \texttt{eyebrow}, \texttt{lip}, \texttt{mouth}, \texttt{nose}, and \texttt{skin}) are determined using formal psychophysical testing. FFSC supports two fine-grained semantics-oriented testing protocols: 1) generalization to novel manipulation methods for the same face attribute, and 2) generalization to novel face attributes.

Moreover, we introduce a new semantics-oriented (SO) face forgery detection method. We first compute the joint probability distribution over the label hierarchy as a way of encoding label relations. Next, we derive marginal probabilities for all labels, each corresponding to a standard binary classification task. This leads to a multi-task learning setting, in which we prioritize the primary task---detecting whether a face image is real or fake---through bi-level optimization~\cite{dempe2002foundations}. Our SO-detection method offers two significant advantages. First, it encourages learning transferable features across manipulations that alter the same face attribute, rather than relying solely on manipulation-specific cues.  Second, it enables integration of features at the semantic level (by detecting manipulations of global face attributes) with those at the signal level (by detecting manipulations in specific face regions) through end-to-end optimization.

Extensive experiments show that the proposed FFSC dataset poses a challenge to current face forgery detectors as a test set and is more effective in inducing more generalizable SO-detectors as a training set, surpassing those trained on FF++~\cite{rossler2019faceforensics}, DFDC~\cite{Dolhansky2020deepfake}, and ForgeryNet~\cite{he2021forgerynet}. Additionally, we demonstrate the superiority of the proposed SO-detection method over the binary and multi-class classification-based counterparts. In summary, our contributions include
\begin{itemize}
	 \item a new definition of face forgery that emphasizes the importance of face semantics,
      \item a new dataset of face forgery that includes a semantic label hierarchy for each image, and 
      \item a new face forgery detection method that is oriented to relying on face semantics.
\end{itemize}

\section{Related Work} \label{sec: related_work}

In this section, we provide a concise review of representative face forgery datasets and detection methods.

\subsection{Face Forgery Datasets}
Four major face manipulation techniques are adopted in the construction of existing face forgery datasets. The first is face editing supplied by Adobe\textsuperscript{\textregistered} Photoshop\textsuperscript{\textregistered} (\eg, Face-Aware Liquify), which provides high-level semantic abstractions for face manipulations~\cite{wang2019detecting}. The second is face swapping based on autoencoders~\cite{faceswap}, which replaces the face from a target image/video with that from the source. The third is the application of conditional generative models~\cite{song2019generative,richardson2018gans}, such as generative adversarial networks (GANs) \cite{goodfellow2014generative} and denoising diffusion models~\cite{song2019generative} for controllable face editing. The fourth is image-based face rendering~\cite{thies2016face2face}, which estimates (morphable) 3D face models from input images/videos \cite{vlasic2005face}, followed by alignment and re-rendering.
Representative face forgery datasets include UADFV~\cite{yang2019exposing}, FaceForensics++ (FF++)~\cite{rossler2019faceforensics}, Celeb-DF~\cite{li2020celeb}, DeepFakeDetection (Google-DFD)~\cite{googledf}, DeeperForensics-1.0 (DF-1.0)~\cite{jiang2020deeperforensics}, DeepFake Detection Challenge (DFDC)~\cite{Dolhansky2020deepfake}, FFIW~\cite{Zhou_2021_CVPR}, ForgeryNet~\cite{he2021forgerynet}, KoDF~\cite{kwon2021kodf}, DF-Platter~\cite{narayan2023df}, and OW-DFA~\cite{Sun_2023_ICCV} with consistent improvements in dataset size (ranging from hundreds~\cite{yang2019exposing} to tens of thousands~\cite{he2021forgerynet}), sample complexity (shifting from single-person to multi-person face forgery~\cite{Zhou_2021_CVPR, narayan2023df}), identity diversity (growing from tens~\cite{yang2019exposing, li2020celeb} to  thousands~\cite{he2021forgerynet}), visual quality (improving from noticeable artifacts to photorealistic outputs), task complexity (evolving from binary classification~\cite{rossler2019faceforensics}, $C$-way classification~\cite{Sun_2023_ICCV, he2021forgerynet,yan2024df40} to forgery localization~\cite{Zhou_2021_CVPR, he2021forgerynet}), and ethical approval.
Table~\ref{tab: datasets_info} presents a summary of these datasets, following the binary and $C$-way classification formulations. In contrast, the proposed FFSC dataset contextualizes face manipulations from the perspective of face semantics.

\begin{table}
  \caption{Summary of existing face forgery datasets}
  \vspace{-.1cm}
  \label{tab: datasets_info}
  \centering
  \small
  \renewcommand\tabcolsep{2pt}
  \resizebox{1\linewidth}{!}{
  \begin{tabular}{lcccc}
    \toprule
    Dataset & \#\,Real Samples & \#\,Fake Samples & \#\,Manipulations & Formulation \\
    \toprule
    UADFV~\cite{yang2019exposing} & 49 & 49 & 1 &  Binary\\
    FF++~\cite{rossler2019faceforensics} & 1,000 & 4,000 & 4  & Binary\\
    Google-DFD~\cite{googledf} & 363 & 3,068 & 5  & Binary\\ 
    Celeb-DF~\cite{li2020celeb} & 590 & 5,639 & 1  & Binary\\
    DF-1.0~\cite{jiang2020deeperforensics} & 50,000 & 10,000 & 1  & Binary\\
    DFDC~\cite{Dolhansky2020deepfake} & 23,564 & 104,500 & 8  & Binary\\
    FFIW~\cite{Zhou_2021_CVPR} & 10,000 & 10,000 & 3  &  Binary \\
    ForgeryNet~\cite{he2021forgerynet} & 99,630 & 121,617 & 15  &  Binary/$C$-way \\
    KoDF~\cite{kwon2021kodf} & 62,166 & 62,166 & 6 & Binary \\
    DF-Platter~\cite{narayan2023df} & 66,630 & 66,630 & 3 & Binary \\
    OW-DFA~\cite{Sun_2023_ICCV} & 25,000 & 34,000 & 20  & $C$-way \\
    DF40~\cite{yan2024df40} & -- & 1M+ & 40 & Binary/$C$-way \\
    \toprule
    FFSC (Ours) & 63,344 & 83,840 & 12  & Semantics-oriented \\
    \bottomrule
  \end{tabular}}

\end{table}

\subsection{Face Forgery Detectors}
Traditional forensics tools are designed to detect image forgeries by spotting statistical irregularities~\cite{popescu2005exposing, popescu2004statistical}, visual artifacts (such as demosaicking~\cite{popescu2005exposing2}, chromatic aberration~\cite{johnson2006exposing}, vignetting~\cite{lyu2010estimating}, and noise~\cite{lyu2014exposing}), and physical and geometric inconsistencies~\cite{johnson2007exposing, kee2014exposing, o2012exposing, conotter2011exposing}.
Face forgery detection has been significantly influenced by these methods as a subfield of forensic science. 
Generally, these knowledge-driven methods are limited by the expressiveness of handcrafted (and manipulation-specific) features.

With the advancements of deep learning, many methods learn to expose face forgery from physiological signals, including eye blinking \cite{Li2018in}, head pose \cite{yang2019exposing}, pupil shape~\cite{guo2022icassp_eyes}, corneal specularity~\cite{hu2021exposing}, and behavioral patterns~\cite{agarwal2019protecting}. Pure data-driven approaches in the spatial~\cite{rossler2019faceforensics, afchar2018mesonet, Li_2019_CVPR_Workshops, nguyen2019capsule, chai2020makes, li2020face, shiohara2022detecting, Xu_2023_ICCV} and frequency~\cite{qian2020thinking,liu2021spatial,luo2021generalizing} domain have also been proposed, coupling with advanced learning strategies, such as attention learning~\cite{zhao2021multi, lu2023detection, yin2023dynamic}, adversarial learning~\cite{chen2022self}, meta-learning~\cite{chen2022ost}, graph learning~\cite{wang2023dynamic, yang2023masked}, contrastive learning~\cite{sun2022dual, Luo2024tifs_CFM, zhang2025learning,choi2024exploiting}, and multi-task learning~\cite{Nguyen2019multi, Sun_2023_ICCV, Cao_2022_CVPR, Yan_2023_ICCV, nguyen2024laa}.
Empirically, data-driven methods tend to overfit training manipulations and struggle to generalize to novel manipulations.

Only until recently have researchers begun detecting fake faces through analysis of face semantics~\cite{haliassos2021lips, dong2022protecting}. Haliassos \etal~\cite{haliassos2021lips} leveraged lipreading features, while Dong \etal~\cite{dong2022protecting} exploited face identity features. In this paper, we further explore this direction, and conduct a systematic reexamination of face forgery at the semantic level. The resulting SO-detectors can be loosely seen as a generalization of the above two methods, with significantly improved detection performance.

\begin{figure*}[!ht]
  \centering
  \subfloat[\texttt{Age} by diffusion autoencoders]{\includegraphics[width=0.247\linewidth]{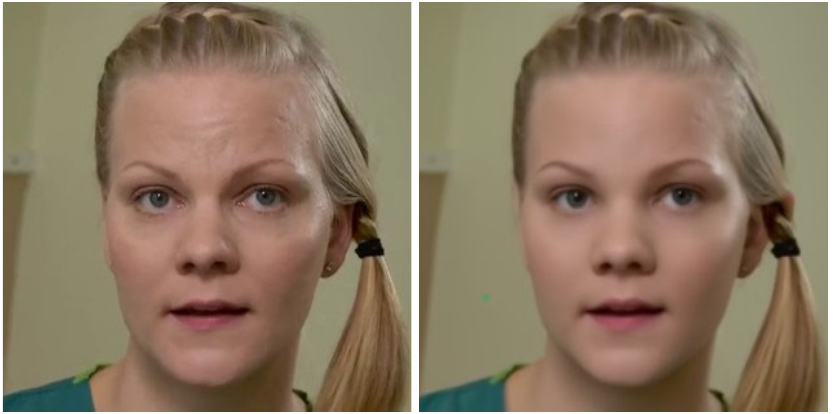}} \hskip.2em
  \subfloat[\texttt{Age} by StyleRes]{\includegraphics[width=0.247\linewidth]{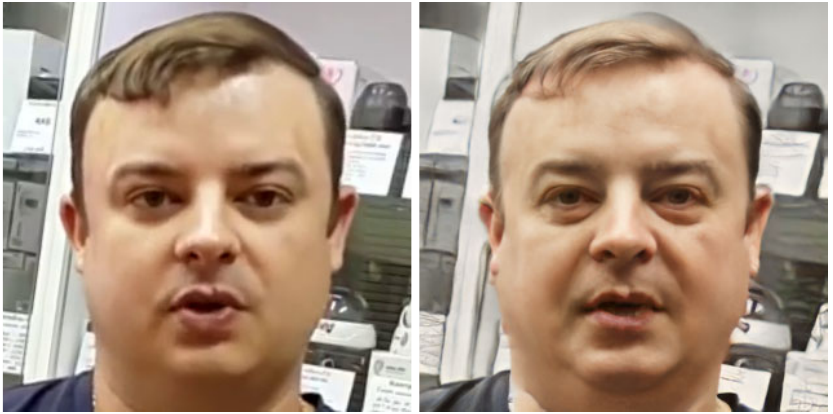}} \hskip.2em
  \subfloat[\texttt{Expression} by first-order motion]{\includegraphics[width=0.247\linewidth]{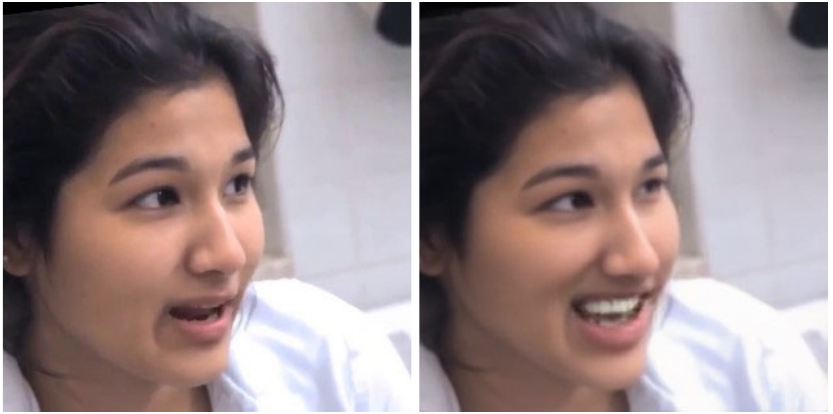}} \hskip.2em
  \subfloat[\texttt{Expression} by StyleRes]{\includegraphics[width=0.247\linewidth]{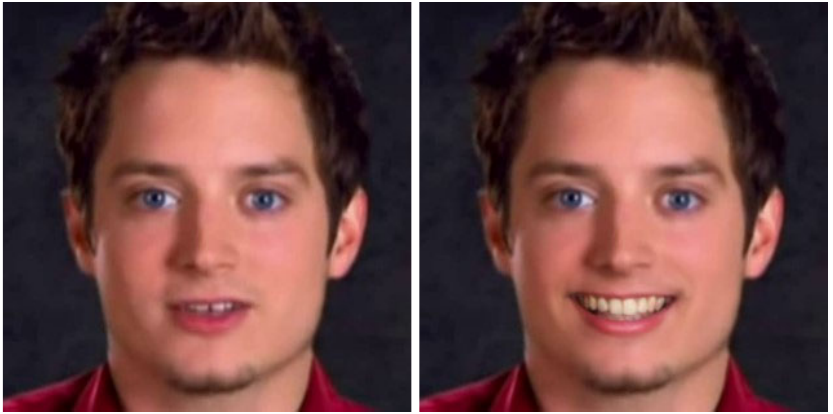}}
  \\
  \subfloat[\texttt{Expression} by first-order motion]{\includegraphics[width=0.247\linewidth]{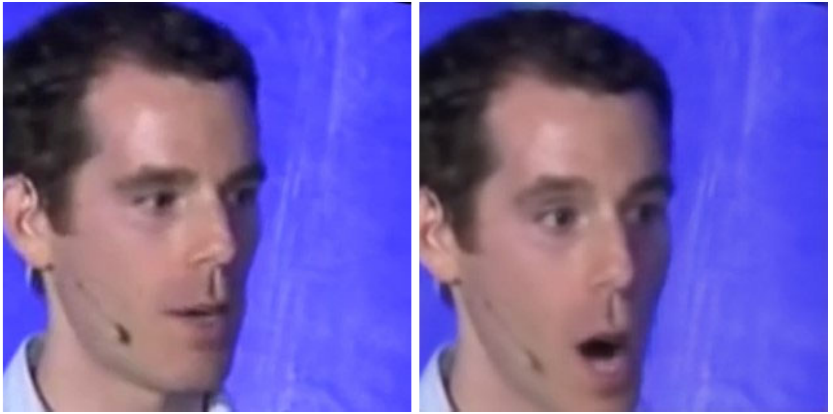}} \hskip.2em
  \subfloat[\texttt{Gender} by StyleGAN2 distillation]{\includegraphics[width=0.247\linewidth]{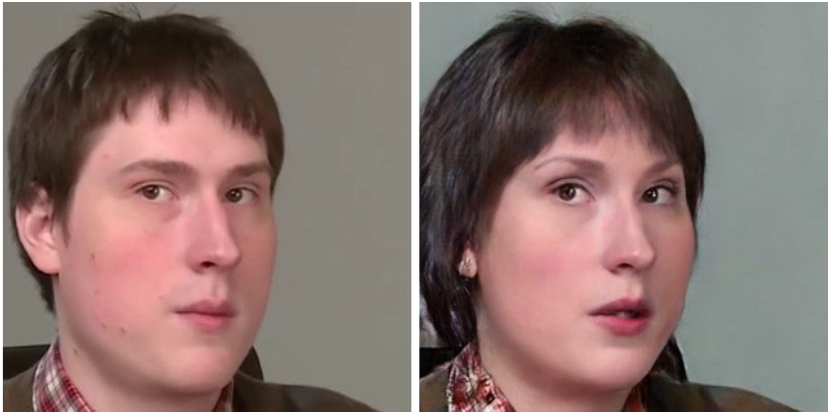}} \hskip.2em
  \subfloat[\texttt{Gender} by diffusion autoencoders]{\includegraphics[width=0.247\linewidth]{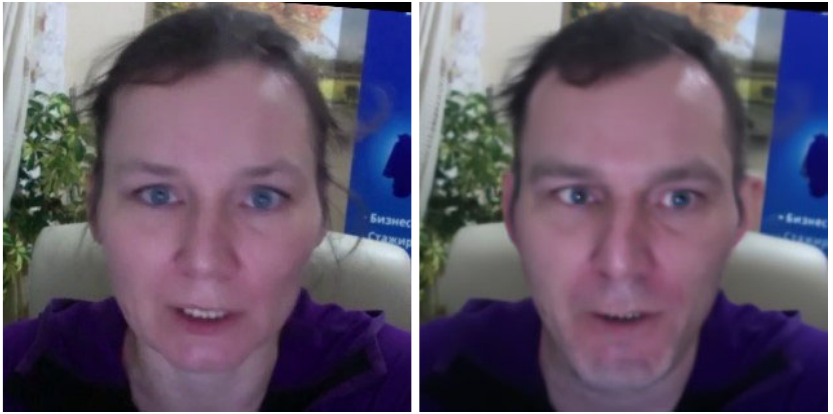}} \hskip.2em
  \subfloat[\texttt{Identity} by SimSwap]{\includegraphics[width=0.247\linewidth]{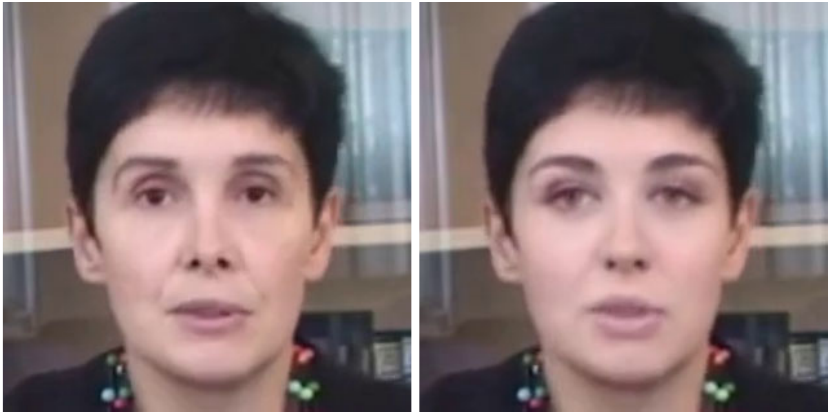}}
  \\
  \subfloat[\texttt{Identity} by FSGAN]{\includegraphics[width=0.247\linewidth]{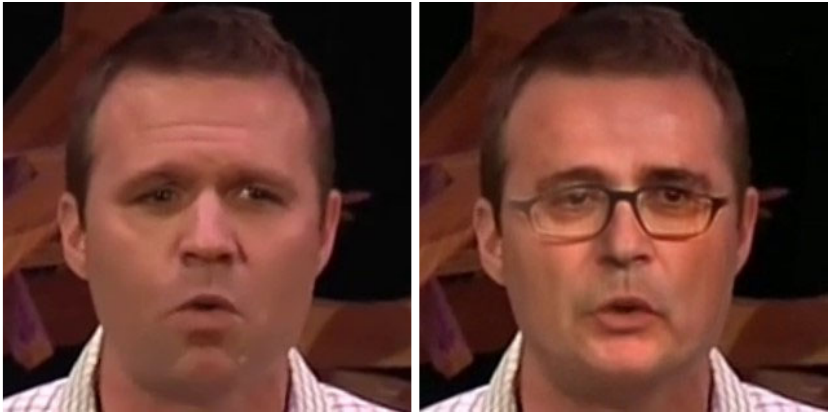}} \hskip.2em
  \subfloat[\texttt{Identity} by BlendFace]{\includegraphics[width=0.247\linewidth]{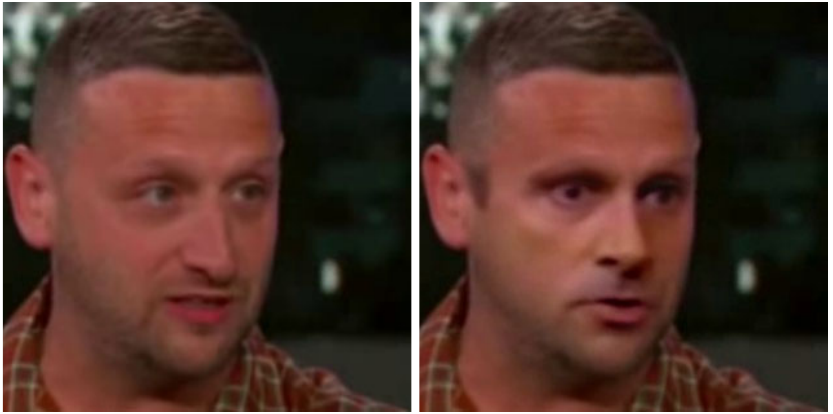}} \hskip.2em
  \subfloat[\texttt{Pose} by first-order motion]{\includegraphics[width=0.247\linewidth]{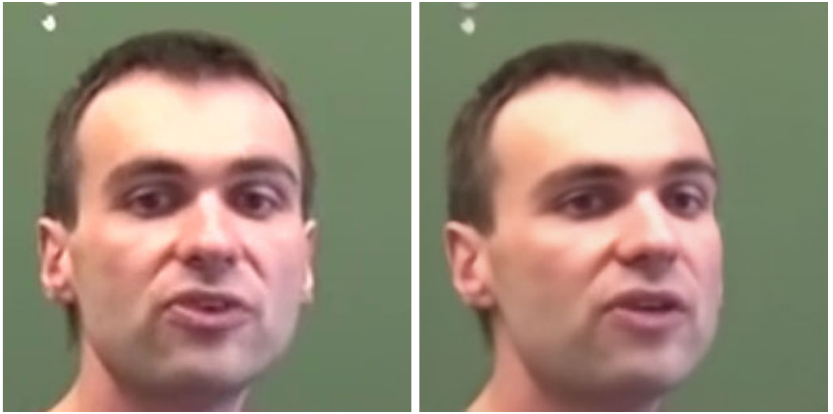}} \hskip.2em
  \subfloat[\texttt{Pose} by thin-plate spline motion]{\includegraphics[width=0.247\linewidth]{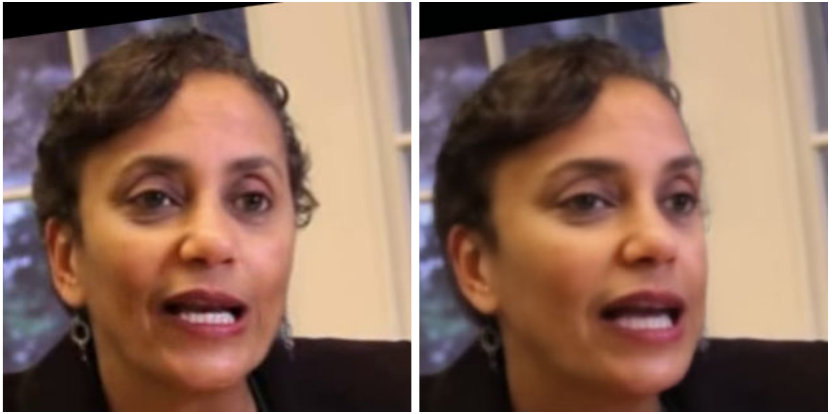}}
  \caption{Face manipulation techniques adopted in FFSC. Each subfigure displays the original and manipulated face images on the left and right, respectively. 
  }
  \label{fig: ffsc_samples}

\end{figure*}

\section{FFSC Dataset} \label{sec: FFSC}

In this section, we introduce the FFSC dataset, including data collection, face manipulation, and data annotation.

\subsection{Data Collection}
We start by collecting real photographic videos from two popular datasets, AVSpeech~\cite{ephrat2018looking} and Celeb-DF YouTube-real~\cite{li2020celeb},
which contain a diverse set of video clips of different ages, expressions, genders, identities, poses, ethnicities, and shooting conditions. 
We randomly select $700$  and $300$ high-resolution videos from the AVSpeech test set and 
Celeb-DF YouTube-real\footnote{Celeb-DF Youtube-real is an additional subset for extended use, and does not overlap with the official Celeb-DF dataset.}, respectively. As most face manipulation methods are image-based, we first uniformly sample $128$ frames from each video and detect the face regions in each frame by RetinaFace~\cite{deng2020retinaface}. We only retain the largest face and extend it to $1.3^2$ times the area of the tight crop produced by the face detector~\cite{rossler2019faceforensics}.
We adjust each face image to a fixed size of $317\times 317$.
Importantly, we exclude face images with low visual quality, closed eyes, extreme poses, and occlusions. This meticulous screening process, conducted by the first author and a research assistant at the City University of Hong Kong, takes approximately one week to complete. In total, we collect $63,344$ real face images, corresponding to $1,000$ face identities.

\subsection{Face Manipulation}\label{subsubsec:fm}
We construct the FFSC dataset by instantiating five global face attribute nodes: \texttt{age}, \texttt{expression}, \texttt{gender}, \texttt{identity}, and \texttt{pose}, through twelve face manipulation methods. 
We leverage eight methods~\cite{Preechakul_2022_CVPR, Pehlivan_2023_CVPR, First_Order_Motion_nips, viazovetskyi2020stylegan2, chen2020simswap, nirkin2019fsgan, li2020face, zhao2022thin} to build the main FFSC dataset, including the well-split training, validation, and test sets with a ratio of $7.8:1.1:1.1$. The independence of face identity is ensured during splitting. We reserve four face manipulation algorithms~\cite{wang2021high, Patashnik_2021_ICCV, Gao_2021_CVPR, zeng2022fnevr} to create an additional test set. This is designed to assess the generalizability of face forgery detectors when exposed to new manipulations that alter the same face attributes in the training set.
To balance between real and fake face images in the main FFSC dataset, we randomly select one-eighth of real images for face manipulation, totaling $75,176$ fake images. As for the additional FFSC test set, we manipulate the real images from the validation and test sets of the main FFSC dataset to obtain $8,664$ fake images.

\noindent \textbf{Age Manipulation.} 
To alter the \texttt{age} attribute, we use two distinct methods, diffusion autoencoders~\cite{Preechakul_2022_CVPR} and StyleRes~\cite{Pehlivan_2023_CVPR}, which are based on denoising diffusion models~\cite{song2020denoising} and GANs for style transfer~\cite{karras2020analyzing}, respectively. These techniques are capable of adjusting the perceived age, either to appear more youthful (see Fig.~\ref{fig: ffsc_samples}(a)) or more aged (see Fig.~\ref{fig: ffsc_samples}(b)).

\noindent \textbf{Expression Manipulation.} 
To alter the \texttt{expression} attribute, we consider two target expressions - smile and surprise (from the neutral face), and adopt two different methods: the first-order motion model~\cite{First_Order_Motion_nips} and StyleRes~\cite{Pehlivan_2023_CVPR}. Unlike StyleRes, which takes a single face image as input, the first-order motion model requires an additional driving video to transform the source face image into a video clip that imitates the face expression in the driving video.

\noindent \textbf{Gender Manipulation.} 
To flip the \texttt{gender} attribute, we adopt diffusion autoencoders~\cite{Preechakul_2022_CVPR} and StyleGAN2 distillation~\cite{viazovetskyi2020stylegan2}, converting 
males to females (see Fig.~\ref{fig: ffsc_samples}(f)) and vice versa (see Fig.~\ref{fig: ffsc_samples}(g)). 
Note that StyleGAN2 distillation is an image-to-image translation method that does not offer adjustable parameters to control the degree of manipulation.

\noindent \textbf{Identity Manipulation.} 
To alter the \texttt{identity} attribute, we follow the practice in existing face forgery datasets~\cite{rossler2019faceforensics, li2020celeb, Dolhansky2020deepfake, he2021forgerynet}, which swaps two faces with different identities by  data-driven SimSwap~\cite{chen2020simswap} and FSGAN~\cite{nirkin2019fsgan}, and knowledge-driven BlendFace~\cite{li2020face}. For each target face, we first search for the most suitable source face by minimizing the Euclidean distance between detected face landmarks~\cite{li2020face}, while excluding faces with the same identity and different gender. Figs.~\ref{fig: ffsc_samples}(h)-(j) show the visual examples, which involve no manipulation degree tuning.

\noindent \textbf{Pose Manipulation.} 
To alter the \texttt{pose} attribute, particularly the head posture in the horizontal plane, we adopt two methods: the first-order motion model~\cite{First_Order_Motion_nips} and the thin-plate spline motion model~\cite{zhao2022thin}. Similar to the former, the spline motion model also requires a driving video as input and allows more complex nonlinear motion transfer. The visual examples are shown in Figs.~\ref{fig: ffsc_samples}(k) and (l).

As for the additional test set in FFSC, we employ four different face manipulation algorithms: HFGI~\cite{wang2021high} (for both \texttt{age} and \texttt{expression} attributes), StyleCLIP~\cite{Patashnik_2021_ICCV} (for the \texttt{gender} attribute), InfoSwap~\cite{Gao_2021_CVPR} (for the \texttt{identity} attribute), and FNeVR~\cite{zeng2022fnevr} (for the \texttt{pose} attribute). HFGI, StyleCLIP, and InfoSwap are GAN-based and only need a single face image as input, whereas FNeVR relies on motion transfer, requiring an additional driving video as input.

\subsection{Data Annotation}\label{subsec: human_label}
\subsubsection{Specification of Manipulation Degree Parameters}
For the face manipulation under consideration, we tune its manipulation degree parameter (if any) to generate a series of manipulated images. We then estimate human discrimination thresholds using a standardized psychophysical procedure~\cite{cornsweet1962staircase,rose1970statistical}. On each trial, three subjects are shown a real face image as the reference and a corresponding manipulated image as the stimulus (for one second and in randomized spatial order), and then asked to indicate whether the two images are perceptually different. This procedure is repeated for $300$ trials for each manipulation degree value and over $50$ image pairs, with ordering determined by a standard staircase method~\cite{cornsweet1962staircase, rose1970statistical}.
The distribution of human responses, as a function of the manipulation degree parameter, is fitted using a Gaussian cumulative distribution function, and the human discrimination threshold is set to the parameter value such that the subjects can clearly perceive the visual differences between the two images $75\%$ of the time~\cite{rose1970statistical}. Finally, we transfer the fitted degree parameter of each manipulation method (on the $50$ real images) to all remaining real images.

\begin{figure*}[t]
  \centering
  \subfloat[\texttt{Age} by diffusion autoencoders]{\includegraphics[width=0.25\linewidth]{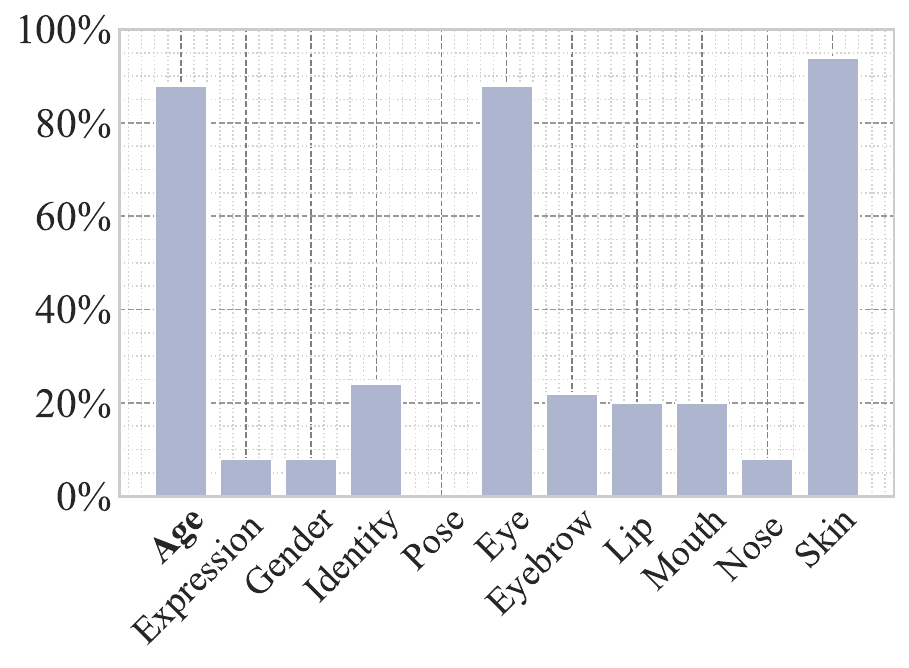}}
  \subfloat[\texttt{Age} by StyleRes]{\includegraphics[width=0.25\linewidth]{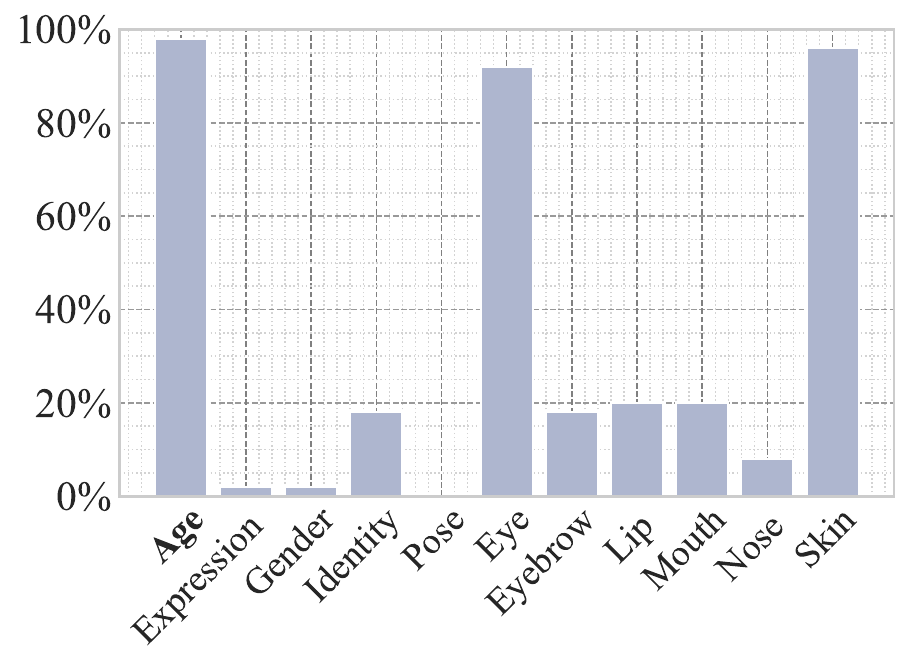}}
  \subfloat[\texttt{Expression} by first-order motion]{\includegraphics[width=0.25\linewidth]{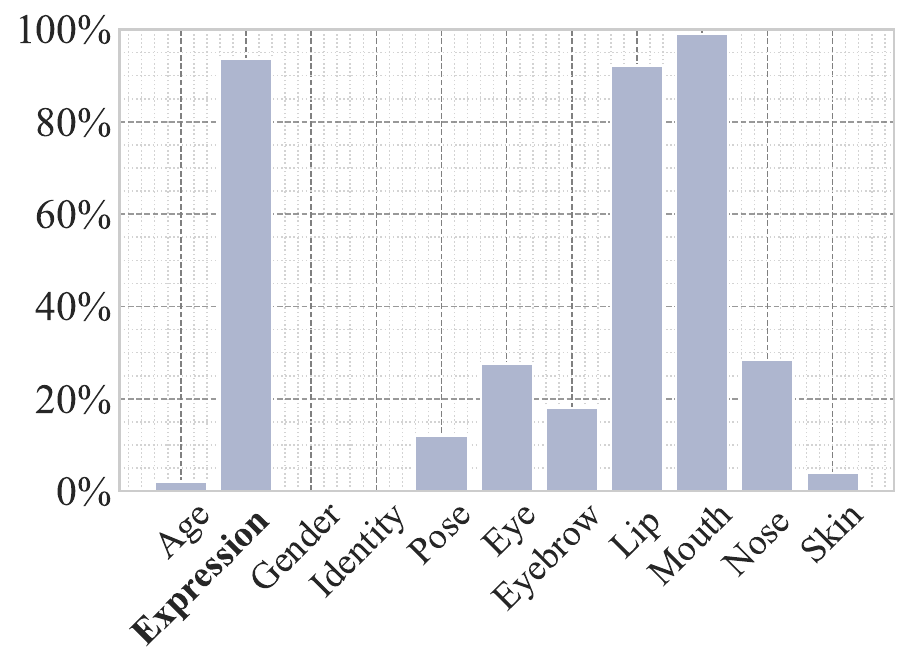}}
  \subfloat[\texttt{Expression} by StyleRes]{\includegraphics[width=0.25\linewidth]{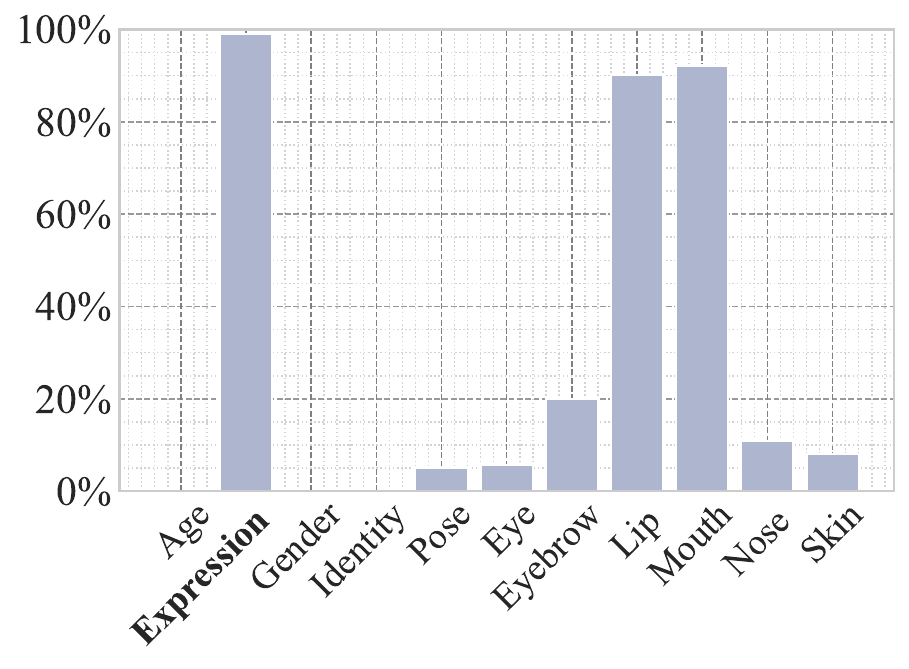}} \\
  
  \subfloat[\texttt{Expression} by first-order motion]{\includegraphics[width=0.25\linewidth]{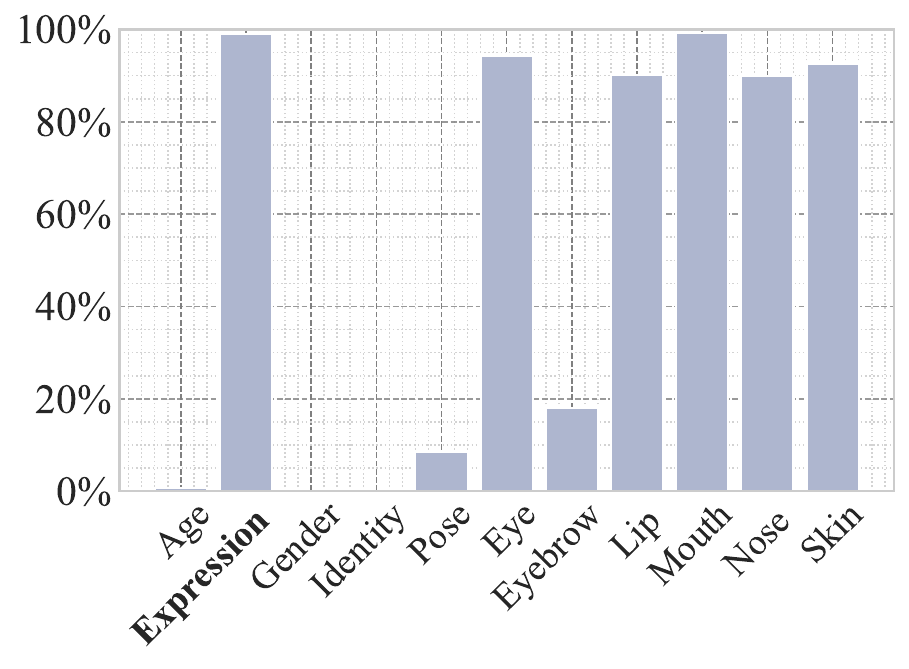}}
  \subfloat[\texttt{Gender} by StyleGAN2]{\includegraphics[width=0.25\linewidth]{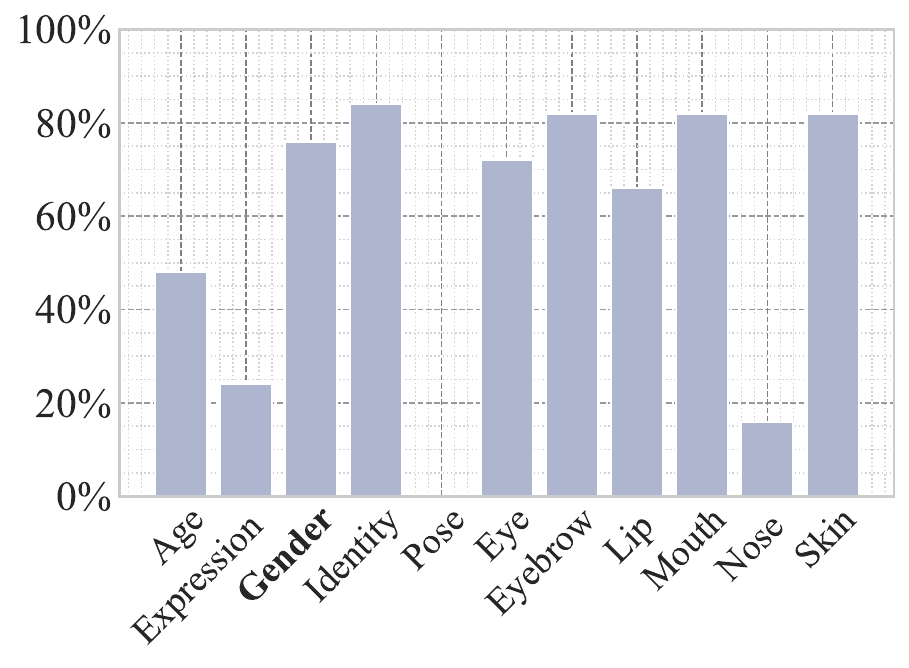}}
  \subfloat[\texttt{Gender} by diffusion autoencoders]{\includegraphics[width=0.25\linewidth]{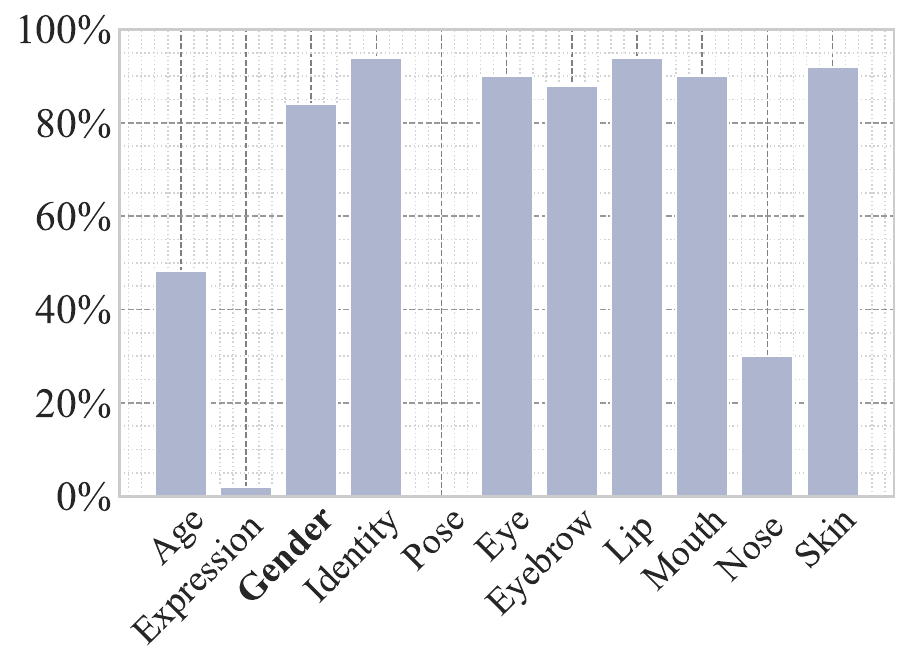}}
  \subfloat[\texttt{Identity} by SimSwap]{\includegraphics[width=0.25\linewidth]{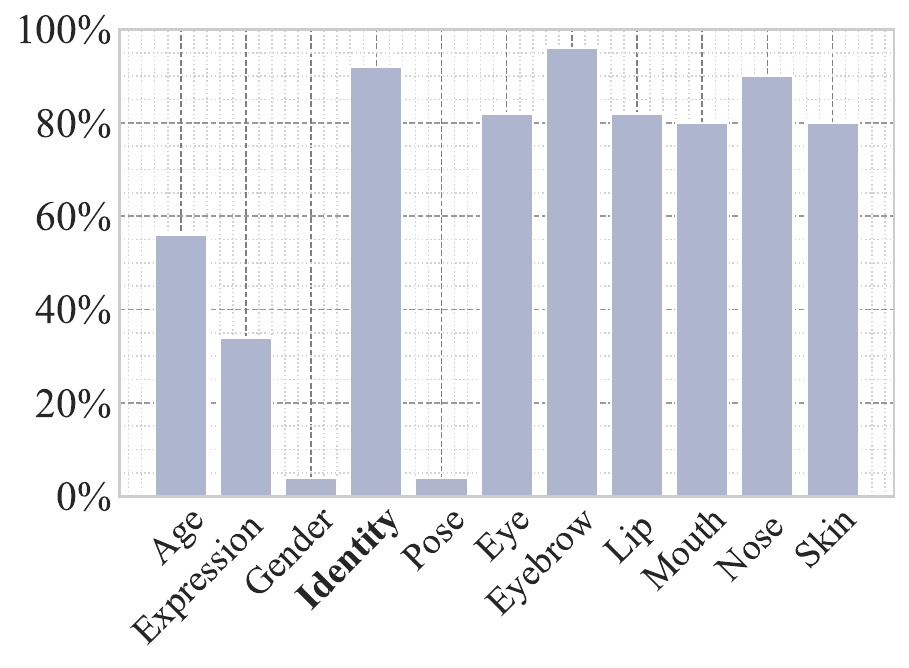}} \\

  \subfloat[\texttt{Identity} by FSGAN]{\includegraphics[width=0.25\linewidth]{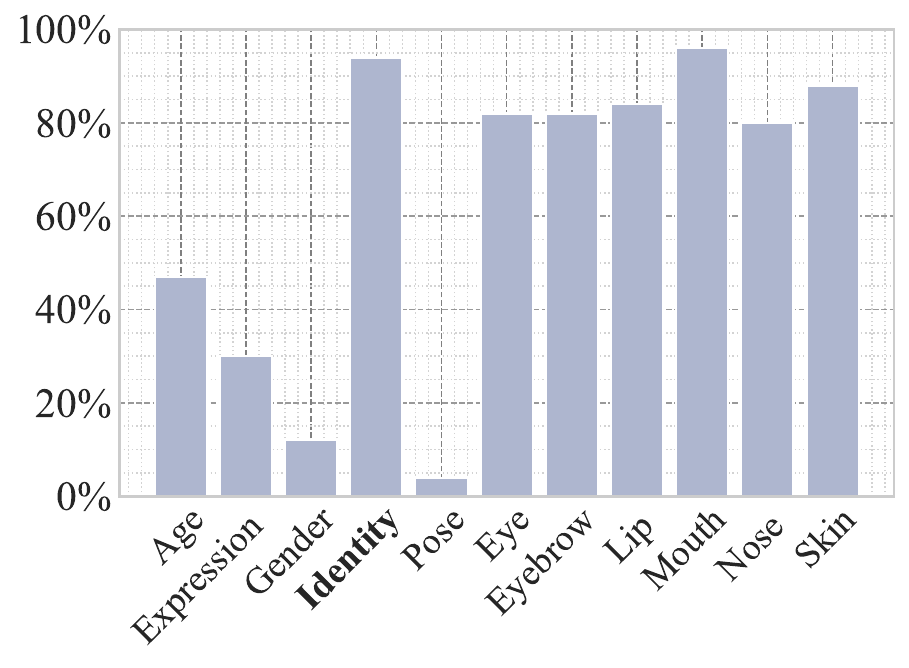}}
  \subfloat[\texttt{Identity} by BlendFace]{\includegraphics[width=0.25\linewidth]{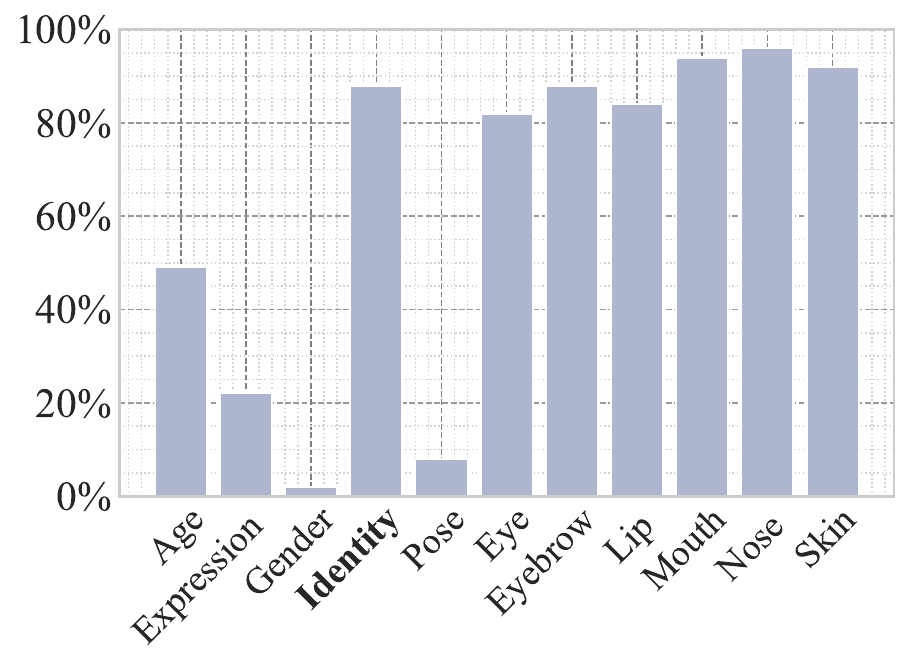}}
  \subfloat[\texttt{Pose} by first-order motion]{\includegraphics[width=0.25\linewidth]{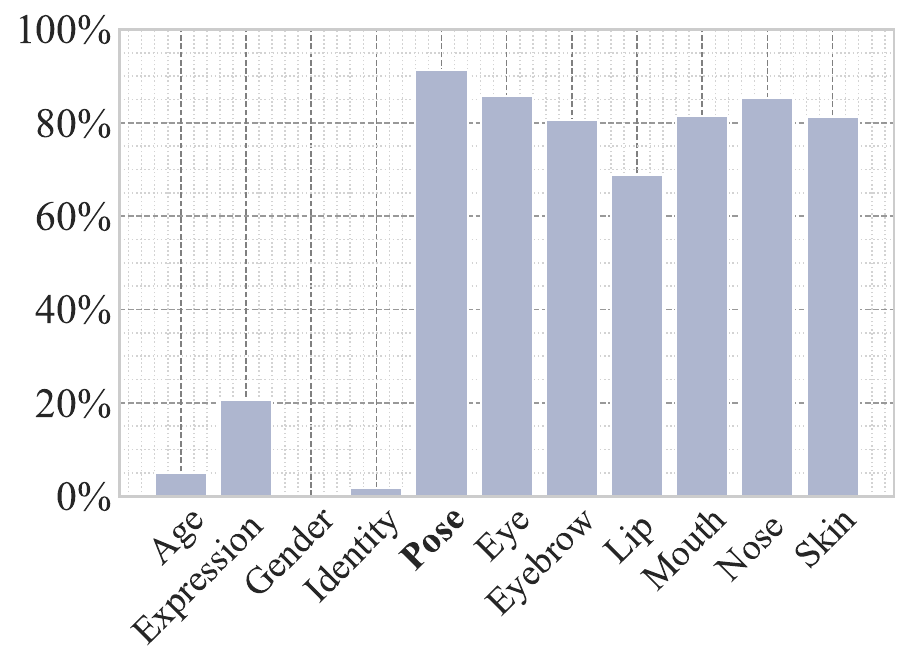}}
  \subfloat[\texttt{Pose} by thin-plate spline motion]{\includegraphics[width=0.25\linewidth]{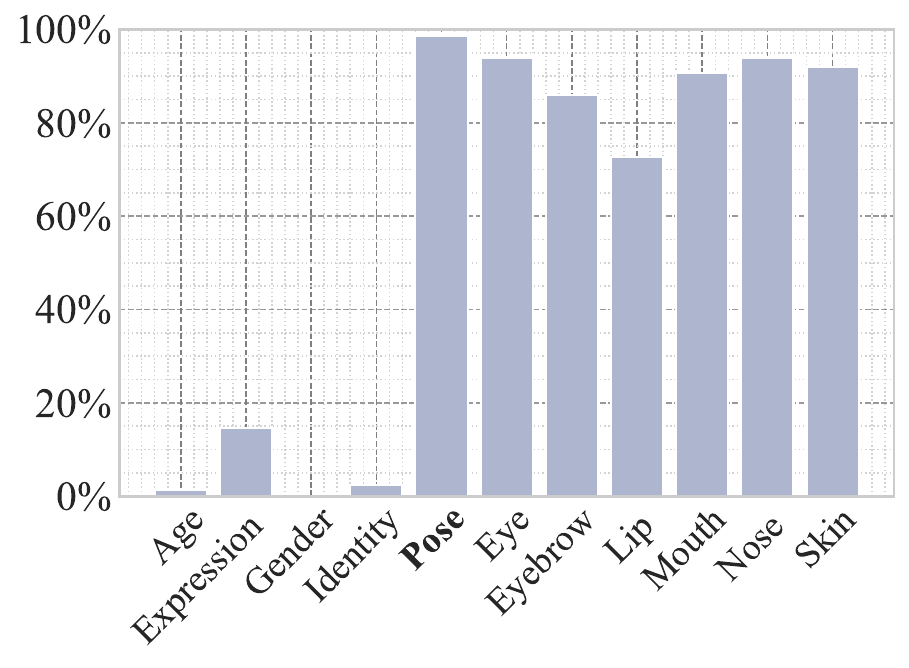}}
  
  \caption{Human discrimination distributions of face attributes and regions. Zoom in for improved visibility.}
  \label{fig: human_labels}
\end{figure*}

\subsubsection{Specification of Label Hierarchy} 
To pair each fake image with a label hierarchy (refer to Fig.~\ref{fig_DAG}(a)), it is essential to determine 1) whether altering one global face attribute impacts other non-targeted attributes and 2) which specific changes in  
local face regions result in the modification of the global face attribute (\ie, the connectivity between the global face attribute to local face regions). Toward this goal, we invite a group of $12$ human subjects to participate in another formal psychophysical experiment. Similarly, a pair of real and manipulated images are shown to at least three subjects (for unlimited time and in randomized spatial order) to indicate whether the modification of non-targeted face attributes (and regions) is visually discriminable, each corresponding to a yes-no task. This process is repeated over $200$ image pairs for each face manipulation method, with human discrimination distributions shown in Fig.~\ref{fig: human_labels}. We find that subjects are generally confident in perceiving the changes in non-targeted face attributes as clearly discriminable (or indiscriminable) with probabilities higher than $70\%$ (or lower than $30\%$), corresponding to a binary label of one (or zero). As for the non-targeted attributes where the probabilities of discrimination fall within the range of $30\%$ to $70\%$---for instance, the \texttt{age} attribute when adjusting for \texttt{gender} and \texttt{identity}---we refrain from assigning definitive binary labels. In such cases, we consider the underlying label of the non-targeted attribute to be \textit{unobserved}. Additionally, the clear discrimination of targeted face attributes, each with a probability over $75\%$, verifies the specification of manipulation degree parameters in the preceding psychophysical experiment.  

The psychophysical results suggest a hierarchical acyclic graph instantiation to encode the semantic labels as shown in Fig.~\ref{fig_DAG}(a). 
Starting from the root node \texttt{face}, we split it into five global face attribute nodes: \texttt{age}, \texttt{expression}, \texttt{gender}, \texttt{identity}, and \texttt{pose}, and connect them to a set of leaf nodes representing non-overlapping local face regions: \texttt{eye}, \texttt{eyebrow}, \texttt{lip}, \texttt{mouth}, \texttt{nose}, and \texttt{skin}. We share the label hierarchy $\bm y \in\{0, 1\}^{N}$, for images with the targeted face attribute manipulated by the same method.  $y_0 = 1$ indicates that the test face image is fake, and vice versa. Likewise, $y_i = 1$, for $i > 1$, indicates the $i$-th face attribute or region is fake, and $y_i= 0$ otherwise. 

The proposed SO-contextualization for face manipulations allows a single manipulation to alter multiple face attributes (\eg, \texttt{age} and \texttt{expression} by StyleRes~\cite{Pehlivan_2023_CVPR}). Meanwhile, different manipulations can affect the same attribute (\eg, \texttt{expression} by FOMM~\cite{First_Order_Motion_nips} and StyleRes). This contextualization discourages the detectors from relying on ``shortcuts'' as observed in~\cite{yan2024df40}, and instead facilitates the learning of more generalizable features.

\section{Semantics-Oriented Face Forgery Detection} \label{sec: proabilistic_loss}
In this section, we describe a new SO-detection method, drawing inspiration from the proposed face forgery definition.

\subsection{Probabilistic Formulation}\label{subsec: semantics-oriented_paradigm}
Conditioning on an input face image $\bm x\in\mathbb{R}^{H\times W\times 3}$, we specify an unnormalized probability distribution over the label hierarchy~\cite{deng2014large}:
\begin{align}
    \label{eq:plf_joint}
    \begin{split}
        \tilde{p}(\bm y|\bm x) =& \prod_i e^{f_i(\bm x)\mathbb{I}[y_i=1]}\prod  _{
           j, i\in \mathcal{P}_j
         } \mathbb{I}\left[\left(\sum_{i} y_i,y_j\right)\ne (0,1)\right]\\
        &\prod  _{
           i , j\in \mathcal{C}_i
         } \mathbb{I}\left[\left( y_i,\sum_{j}y_j\right)\ne (1,0)\right],
    \end{split}
\end{align}
where $f_i(\bm x)$ can be regarded as  the raw score (\ie, confidence) of the detector $\bm f$ on the $i$-th label. $\mathbb{I}[\cdot]$ is an indication function that excludes two illegal label relations. If the $j$-th node is manipulated  (\ie, $y_j = 1$), it necessitates that at least one of its parents with indices in $\mathcal{P}_j$ is also manipulated, ensuring that $\sum_i y_i \neq 0$. Conversely, if the $i$-th node is manipulated, at least one of its children with indices in $\mathcal{C}_i$ should also be manipulated, leading to $\sum_j y_j \neq 0$.

We then normalize Eq.~\eqref{eq:plf_joint} to obtain the joint probability 
\begin{align}\label{eq:joint}
    p(\bm y|\bm x)=\frac{\tilde{p}(\bm y|\bm x)}{Z(\bm x)}, \,\mathrm{where}\, Z(\bm x) = \sum_{\bm y'} \tilde{p}(\bm y'|\bm x).
\end{align}
Given that our label hierarchy is not complex (see Fig.~\ref{fig_DAG}(a)), $Z(\bm x)$ can be computed exhaustively via matrix multiplication.

\subsection{Bi-level Optimization}
Assume a training minibatch $\mathcal{B}_\mathrm{tr} = \{\bm x^{(k)}, \bm y^{(k)}, \mathcal{I}^{(k)}\}_{k=1}^K$, where $\bm y ^{(k)} = \{0,1\}^N$ is the complete ground truth vector and $\mathcal{I}^{(k)}\subseteq \{0, 1,\ldots, N-1\}$ is the index set of observed labels. Our goal is to optimize a differentiable function $\bm f(\bm x; \bm \theta)$, parameterized by a vector $\bm \theta$, that outputs the raw scores of all $N$ labels, based on which the joint probability $p(\bm y|\bm x;\bm \theta)$ in Eq.~\eqref{eq:joint} can be computed. A straightforward optimization objective is to minimize the negative log joint likelihood of the observed labels:
\begin{align}
    \label{eq: plf}
        \ell_\mathrm{JL}(\mathcal{B}_\mathrm{tr};\bm \theta) = & -\frac{1}{K}\sum_{k=1}^{K} \log p(\bm y^{(k)}_{\mathcal{I}^{(k)}}\vert \bm x^{(k)};\bm \theta) \nonumber\\
        = &-\frac{1}{K}\sum_{k=1}^{K} \log \!\! \sum_{\bm y: \bm y_{\mathcal{I}^{(k)}}=\bm y^{(k)}_{\mathcal{I}^{(k)}} }p(\bm y \vert \bm x^{(k)};\bm \theta), 
\end{align}
where we marginalize the unobserved labels (\ie, the second summation in Eq. \eqref{eq: plf}) in computing the joint likelihood~\cite{deng2014large}. However, Eq.~\eqref{eq: plf} does not facilitate prioritization of the primary task, \ie, detecting whether a face image is real or fake. Thus, we opt for the multi-task learning setting and minimize an alternative loss function that represents a linear weighted sum of the negative log marginal likelihoods of the observed labels:
\begin{align}
    \label{eq: w-plf}
        \ell_\mathrm{ML}(\mathcal{B}_\mathrm{tr};\bm \theta)
        =&-\frac{1}{K}\sum_{k=1}^{K} \sum_{i \in \mathcal{I}^{(k)}} \lambda_{i}\log p( y^{(k)}_i|\bm x^{(k)};\bm \theta)\nonumber\\
        =&-\frac{1}{K}\sum_{k=1}^{K} \sum_{i \in \mathcal{I}^{(k)}} \lambda_{i}\log \sum_{\bm y:  y_{i}=y^{(k)}_{i} }p(\bm y \vert \bm x^{(k)};\bm \theta),
\end{align}
where $\bm \lambda = [\lambda_0,
\lambda_1,
\ldots,\lambda_{N-1}]^\intercal$ is the loss weight vector, trading off the $N$ tasks.
To prioritize the primary task and to automate the loss weight adjustment, we resort to bi-level optimization~\cite{dempe2002foundations}, in particular, the Auto-$\lambda$ algorithm~\cite{liu2022auto_lambda}. At the upper level, we only minimize the loss of the primary task with respect to $\bm \lambda$ on the validation minibatch $\mathcal{B}_\mathrm{val}$. At the lower level, we minimize the overall loss defined in Eq.~\eqref{eq: w-plf} with respect to the model parameters $\bm \theta$ on the training minibatch $\mathcal{B}_\mathrm{tr}$. This leads to
\begin{align}
    \label{eq:auto-weighting}
        \min_{\bm \lambda} -& \frac{1}{\vert\mathcal{B}_{\mathrm{val}}\vert } \sum_{\bm x \in\mathcal{B}_{\mathrm{val}}} 
        \log p(y_0|\bm x;\bm \theta^\star)
        \nonumber\\
        \text{s.t.}\quad \bm \theta^\star &= \argmin_{\bm \theta}\ell_\mathrm{ML}(\mathcal{B}_\mathrm{tr};\bm \theta),
\end{align}
where $\vert\cdot\vert$ denotes the cardinality of a set. 
During optimization, we sample training and validation data as different minibatches in the same training dataset. 
Solving Problem~\eqref{eq:auto-weighting} necessitates calculating second-order derivatives, which can be computationally and memory intensive. To address this, we employ the finite difference method as an efficient approximator~\cite{liu2022auto_lambda}.

\begin{table*}[]
\small
\renewcommand{\arraystretch}{1.2}
\centering 
\caption{Results of thirteen face forgery detectors under the testing Protocol-1 and Protocol-2 on the combined main and additional test sets of FFSC. All methods are trained on the training set of FF++ (or its respective augmented versions).  ``\xmark'' indicates that no such evaluation can be properly done.
The top two results are highlighted in bold}
\resizebox{0.9\textwidth}{!}{
\begin{tabular}{l|cc|cccc|cccccc}
\toprule
\multirow{3}{*}{Method} & \multicolumn{2}{c|}{\multirow{2}{*}{Intra-dataset }} & \multicolumn{4}{c|}{Protocol-1 }                  & \multicolumn{6}{c}{Protocol-2 }                                   \\
\cline{4-13}
                        & \multicolumn{2}{c|}{}                               & \multicolumn{2}{c}{\texttt{Expression}} & \multicolumn{2}{c|}{\texttt{Identity}} & \multicolumn{2}{c}{\texttt{Age}} & \multicolumn{2}{c}{\texttt{Gender}} & \multicolumn{2}{c}{\texttt{Pose}} \\
\cline{2-13}
                        & \texttt{Acc}                      & \texttt{AUC}                     & \texttt{Acc}            & \texttt{AUC}           & \texttt{Acc}           & \texttt{AUC}          & \texttt{Acc}        & \texttt{AUC}        & \texttt{Acc}          & \texttt{AUC}         & \texttt{Acc}         & \texttt{AUC}        \\
\hline
Rossler19~\cite{rossler2019faceforensics} & 98.61 & 99.73
& 52.02 & 64.01 & 68.13 & 72.65
& 67.95 & 78.54 & 73.21 & 73.99 & 61.65 & 74.50 \\
CNND~\cite{wang2019cnngenerated} & 98.13 & 99.56                        
& 52.36 & 60.12 & 68.52 & 74.41 
& 64.77 & 75.46 & 68.75 & 80.65 & 58.04 & 72.03 \\
F3-Net~\cite{qian2020thinking} & 97.87 & 99.70
& 57.78 & 71.94 & 66.99 & 75.12 
& 60.65 & 76.31 & 69.66 & 82.48 & \textbf{71.99} & 85.66 \\
FFD~\cite{FFD} & 97.54 & 98.58
& 52.16 & 66.92 & 60.00 & 67.83
& 62.39 & 74.10 & 72.90 & 76.55 & 70.71 & \textbf{90.49} \\
Patch-Forensics~\cite{chai2020makes} & 98.14 & 99.95
& 57.07 & 77.31 & 60.80 & 81.36 
& 54.23 & 68.03 & 60.85 & 67.81 & 53.07 & 70.96 \\
Face X-ray~\cite{li2020face} & 93.48 & 98.37
& 62.90 & \textbf{92.86} & 70.38 & \textbf{93.11}
& \textbf{81.99} & \textbf{97.96} & 65.17 & 81.52 & 66.99 & 82.03 \\
MADD~\cite{zhao2021multi} & 97.68 & 99.51 
& \textbf{74.94} & \textbf{82.22} & \textbf{84.20} & 91.63
& 67.05 & 75.04 & \textbf{83.13} & \textbf{91.48} & 71.59 & 79.72 \\
FRDM~\cite{luo2021generalizing} & 98.32  &  99.41  
& 50.68 & 60.15 & 72.16 & 87.71
& 60.77 & 68.31 & 64.74 & 71.04 & 61.53 & 71.45 \\
Lip-Forensics~\cite{haliassos2021lips} & 97.81 & 98.83
& \xmark & \xmark & 60.00 & 76.76             
& 52.16 & 60.75 & 55.96 & 74.26 & \xmark & \xmark \\
RECCE~\cite{Cao_2022_CVPR} & 97.06 & 99.32    
& 57.78 & 75.33 & 61.08 & 80.77
& 55.11 & 67.46 & 60.26 & 81.76 & 55.74 & 73.61 \\
SBI~\cite{shiohara2022detecting} & 97.33 & 99.64 
& \textbf{73.52} & 79.81 & \textbf{85.52} & \textbf{96.37} 
& \textbf{85.94} & \textbf{92.64} & 70.82 & 75.80 & 61.56 & 68.82 \\
ICT~\cite{dong2022protecting} & 88.20 & 94.39  
& \xmark    & \xmark    & 76.89 & 84.20
& 66.23 & 72.18 & 51.65 & 52.83 & 70.08 & 76.12 \\
CADDM~\cite{Dong_2023_CVPR} & 98.77 & 99.70 
& 65.85 & 73.61 & 78.24 & 87.14
& 78.98 & 85.31 & \textbf{78.86} & \textbf{87.14} & \textbf{78.66} & \textbf{88.17} \\
\bottomrule
\end{tabular}
}
\label{tab: results_benchmarking_SOTA}
\end{table*}

\section{Experiments}
In this section, we first employ the proposed FFSC dataset as the test set for evaluating current face forgery detectors. We next compare FFSC with current face forgery datasets~\cite{rossler2019faceforensics, Dolhansky2020deepfake, he2021forgerynet} in fostering generalization as training sets. Last, we highlight the superiority of our SO-detection method over the traditional binary and multi-class classification-based detectors. 

\subsection{FFSC as the Test Set}~\label{sec: bench_protocols}
We introduce two new testing protocols with the goal of facilitating a fine-grained assessment of face forgery detectors, yielding valuable insights into their relative performance.
\begin{itemize}
    \item \textbf{Protocol-1}: Generalization to novel manipulation methods for the same face attribute in the training set.
    \item \textbf{Protocol-2}: Generalization to novel face attributes absent from the training set.
\end{itemize}

\begin{table}[!t]
\small
\renewcommand{\arraystretch}{1.2}
\centering 
\caption{AUC results of various base detectors retrained on different datasets. Intra-dataset results are omitted, as denoted by ``--''
}
\resizebox{\linewidth}{!}{
\begin{tabular}{llcccc}
\toprule
Base Model                   & Training  & FF++  & DFDC  & DF-1.0 & Celeb-DF   \\
\hline
\multirow{4}{*}{Rossler19~\cite{rossler2019faceforensics}} 
                         & FF++    & --   & 62.19 & 85.41  & 73.70 \\
                         & DFDC    & 71.34 & --   & 79.60  & 71.19 \\
                         & ForgeryNet  & 85.06 & 71.08 & \textbf{90.09}  & 73.06 \\
                         \cline{2-6}
                         & FFSC (Ours) & \textbf{92.99}  & \textbf{77.29} & 89.94  & \textbf{82.23}      \\
\hline
\multirow{4}{*}{CNND~\cite{wang2019cnngenerated}}    
                         & FF++     & --    & 72.10 & 74.40 & 75.60  \\
                         & DFDC     & 73.34 & --    & 69.32 & 64.79     \\
                         & ForgeryNet  & 66.81 & \textbf{75.55} & 76.49 & 69.82     \\
                         \cline{2-6}
                         & FFSC (Ours)   & \textbf{88.01}  & 72.37 & \textbf{92.26}  & \textbf{83.30}       \\
\hline
\multirow{4}{*}{MADD~\cite{zhao2021multi}}    
                         & FF++   & --    & 67.94 & 66.58 & 77.44      \\
                         & DFDC   & 74.22 & --    & 74.32 & 65.53      \\
                         & ForgeryNet  & 82.74 & 74.53 & 82.10 &  73.53     \\
                         \cline{2-6}
                         & FFSC (Ours)  & \textbf{92.70} & \textbf{78.90}  & \textbf{88.67}  & \textbf{88.61}     \\
\hline                         
\multirow{4}{*}{FRDM~\cite{luo2021generalizing}}    
                         & FF++   & --    & \textbf{79.70} & 73.80 &  79.40  \\
                         & DFDC   & 67.33 & --    & 74.11  & 63.80  \\
                         & ForgeryNet & 71.88 & 74.97 & 78.07  & 74.38   \\
                         \cline{2-6}
                         & FFSC (Ours) & \textbf{93.54}  & 76.64 & \textbf{86.48}  & \textbf{83.88}       \\
\hline                         
\multirow{4}{*}{RECCE~\cite{Cao_2022_CVPR}}   
                         & FF++   & --    & 69.06 & 63.05 & 68.71  \\
                         & DFDC   & 61.63 & --    & 69.86 & 63.42  \\
                         & ForgeryNet  & 68.47 & \textbf{75.68} & 72.82  & 74.98  \\
                         \cline{2-6}
                         & FFSC (Ours) & \textbf{97.11} & 71.78 & \textbf{81.36} & \textbf{80.12}       \\
\hline
\multirow{4}{*}{CADDM~\cite{Dong_2023_CVPR}}   
                         & FF++   & --  & 59.22 & 64.27 & 68.28 \\
                         & DFDC  & 61.69 & --  & 73.48 & 63.63 \\
                         & ForgeryNet  & 72.98 & \textbf{75.16}  & 73.75 & 67.44   \\
                         \cline{2-6}
                         & FFSC (Ours)  & \textbf{87.10}   & 71.81 & \textbf{88.08}  & \textbf{83.63}       
\\
\bottomrule
\end{tabular}}
\label{tab: results_differnet_datasets}
\end{table}

We examine thirteen face forgery detectors, including Rossler19~\cite{rossler2019faceforensics}, CNND~\cite{wang2019cnngenerated}, F3-Net~\cite{qian2020thinking}, FFD~\cite{FFD}, Patch-Forensics~\cite{chai2020makes}, Face X-ray~\cite{li2020face}, MADD~\cite{zhao2021multi}, FRDM~\cite{luo2021generalizing}, Lip-Forensics~\cite{haliassos2021lips}, RECCE~\cite{Cao_2022_CVPR}, SBI~\cite{shiohara2022detecting}, ICT~\cite{dong2022protecting}, and CADDM~\cite{Dong_2023_CVPR}. Rossler19, CNND, and Patch-Forensics are commonly compared baseline models. F3-Net and FRDM expose face forgery by high-frequency analysis. Face X-ray and SBI learn to detect blending boundaries. FFD and MADD employ attention mechanisms to extract the most relevant features. RECCE and CADDM, respectively, adopt face reconstruction and manipulation localization as auxiliary tasks. Lip-Forensics and ICT rely primarily on high-level lipreading and face identity features, respectively, rather than signal-level analysis.

We adopt the prediction accuracy (\ie, \texttt{Acc} (\%)) and the area under the receiver operating characteristic curve (\ie, \texttt{AUC} (\%)) as the evaluation metrics. 
The results are shown in Table~\ref{tab: results_benchmarking_SOTA}. Despite the superior intra-dataset performance achieved by nearly all detectors, they struggle to generalize to novel manipulations that alter the same face attribute. This provides a strong indication that existing detectors rely heavily on manipulation-specific features that are less generalizable. Two exceptions are Face X-ray~\cite{li2020face} and SBI~\cite{shiohara2022detecting} that deliver satisfactory results under Protocol-1, due to their ability to detect blending boundaries commonly found in expression and identity manipulations. However, their effectiveness diminishes under Protocol-2, especially for the \texttt{gender} and \texttt{pose} attributes, where blending techniques are not involved during manipulation.
Engineered to identify local visual artifacts, CADDM~\cite{Dong_2023_CVPR}  performs remarkably under Protocol-2, suggesting that current techniques for manipulating \texttt{age}, \texttt{gender}, and \texttt{pose} still have room for improvements in terms of visual fidelity. Correspondingly, CADDM achieves subpar performance for the \texttt{expression} attribute, which focuses on local semantic changes with fewer noticeable artifacts. 

The performance of the semantics-based detectors Lip-Forensics~\cite{haliassos2021lips} and ICT~\cite{dong2022protecting} is not exceptional under either Protocol-1 or Protocol-2. This could be attributed to the fact that these detectors only exploit a single semantic face attribute, and do not capture the intrinsic interactions of multiple face semantics and their relationships with local face regions.

In summary, the assessment results on the proposed FFSC dataset underscore the ongoing challenge in constructing generalizable face forgery detectors across different manipulation methods and face attributes. Presently, detectors are predominantly dependent on cues specific to training manipulations, particularly those accompanied by visual distortions.
Furthermore, our new testing protocols have revealed certain deficiencies in current semantics-based detectors, inspiring us to re-examine the role of face semantics in developing face forgery detectors.

\subsection{FFSC as the Training Set} \label{sec: experiments}

\begin{table*}[!t]
\small
\renewcommand{\arraystretch}{1.2}
\centering 
\caption{Protocol-1 test results. 
All detectors are trained on the training set of FFSC and tested on the additional test set of  FFSC
}
\resizebox{0.74\textwidth}{!}{
\begin{tabular}{rcccccccccc}
\toprule
\multirow{3}{*}{Base Model}   & \multicolumn{10}{c}{Protocol-1} \\
\cline{2-11}
 & \multicolumn{2}{c}{\texttt{Age}} & \multicolumn{2}{c}{\texttt{Expression}} & \multicolumn{2}{c}{\texttt{Gender}} & \multicolumn{2}{c}{\texttt{Identity}} & \multicolumn{2}{c}{\texttt{Pose}} \\
\cline{2-11}
                           & \texttt{Acc}                 & \texttt{AUC}                 & \texttt{Acc}        & \texttt{AUC}        & \texttt{Acc}            & \texttt{AUC}           & \texttt{Acc}          & \texttt{AUC}         & \texttt{Acc}           & \texttt{AUC}                 \\
\hline

Rossler19 & 86.60 & 98.34 & \textbf{88.04} & \textbf{99.13} & 88.24 & 98.70 & 83.48 & 95.49 & 80.31 & 89.14 \\
SO-Rossler19 & \textbf{95.73}      & \textbf{98.87}      & 85.01       & 98.81      & \textbf{95.63}        & \textbf{98.90}        & \textbf{94.78}     & \textbf{97.77}      & \textbf{84.24}      & \textbf{92.11} \\
\hline
     
CNND & 81.29 & 97.17 & 81.25 & \textbf{98.59} & 80.86 & 98.22 & 77.65 &\textbf{92.15} & 70.91 & 80.58 \\
SO-CNND & \textbf{91.61}      & \textbf{97.24}      & \textbf{91.70}        & 97.62      & \textbf{93.21}        & \textbf{98.90}        & \textbf{80.51}     & 86.87      & \textbf{79.18}      & \textbf{84.42} \\
\hline
     
MADD & \textbf{94.06} & 99.20 & \textbf{95.64} & \textbf{99.71} & \textbf{95.45} & 99.58 & 89.94 & 96.52 & \textbf{88.44} & 94.93 \\
SO-MADD & 93.70       & \textbf{99.30}    & 92.89      & 98.93      & 94.18        & \textbf{99.59}       & \textbf{93.09}    & \textbf{98.14}    & 87.80       & \textbf{95.20} \\
\hline
     
FRDM & 83.99 & 93.59 & 84.94 & 93.82 & 85.27 & 95.53 & 71.52 & 80.72 & 77.79 & 85.61 \\
SO-FRDM & \textbf{86.60}       & \textbf{96.41}      & \textbf{86.78}       & \textbf{96.99}      & \textbf{87.07}        & \textbf{98.07}      & \textbf{80.35}     & \textbf{88.22}      & \textbf{78.30}       & \textbf{87.95} \\
\hline
    
RECCE & 94.06 & \textbf{99.34} & 97.48 & \textbf{99.76} & 97.16 & 99.75 & 91.98 & 97.94 & 86.87 & 93.99 \\
SO-RECCE & \textbf{95.94}     & 99.21     & \textbf{98.08}       & 99.20     & \textbf{99.27}       & \textbf{99.92}      & \textbf{97.45}     & \textbf{99.09}     & \textbf{90.12}     & \textbf{96.87} \\
\hline
     
CADDM & 80.44 & \textbf{96.67} & 79.72 & \textbf{97.67} & 81.53 & 97.72 & 74.56 & 89.11 & 74.35 & 85.77 \\
SO-CADDM & \textbf{86.65}      & 96.09      & \textbf{89.25}       & 97.46      & \textbf{86.98}        & \textbf{98.06}       & \textbf{85.43}     & \textbf{90.99}      & \textbf{79.43}      & \textbf{88.20} \\  
\bottomrule
\end{tabular}
}
\label{tab: results_protocol1_retrain}
\end{table*}

\begin{table*}[!t]
\small
\renewcommand{\arraystretch}{1.2}
\centering 
\caption{Protocol-2 test results. All detectors are trained on four out of five face attributes in the training set of FFSC and tested on the remaining face attribute in the main test set of  FFSC 
}
\resizebox{0.74\textwidth}{!}{
\begin{tabular}{rcccccccccc}
\toprule
\multirow{3}{*}{Base Model}    & \multicolumn{10}{c}{Protocol-2}  
\\
\cline{2-11}
                           & \multicolumn{2}{c}{\texttt{Age}} & \multicolumn{2}{c}{\texttt{Expression}} & \multicolumn{2}{c}{\texttt{Gender}} & \multicolumn{2}{c}{\texttt{Identity}} & \multicolumn{2}{c}{\texttt{Pose}} \\
\cline{2-11}
                          & \texttt{Acc}        & \texttt{AUC}        & \texttt{Acc}            & \texttt{AUC}           & \texttt{Acc}          & \texttt{AUC}         & \texttt{Acc}           & \texttt{AUC}          & \texttt{Acc}         & \texttt{AUC}        \\
\hline

Rossler19 & 77.90 & 88.77 & 74.43 & 82.80 & 79.23 & 90.42 & 56.87 & 77.33 & 77.36 & 90.30 \\
SO-Rossler19 & \textbf{83.92}      & \textbf{92.66}      & \textbf{78.66}       & \textbf{88.74}      & \textbf{84.29}        & \textbf{93.30}       & \textbf{64.66}      & \textbf{81.26}     & \textbf{83.30}       & \textbf{92.09} \\
\hline
    
CNND & 70.37 & 78.04 & 68.15 & 77.78 & 70.48 & 81.41 & 57.95 & 63.12 & 70.39 & 79.15  \\
SO-CNND & \textbf{73.86}      & \textbf{82.31}      & \textbf{76.48}       & \textbf{85.44}      & \textbf{78.23}        & \textbf{89.39}       & \textbf{60.34}      & \textbf{70.45}     & \textbf{73.21}       & \textbf{82.36} \\
\hline
    
MADD & 66.36 & 80.53 & 63.61 & 74.78 & 77.95 & \textbf{89.99} & 53.30 & 75.55 & 73.66 & \textbf{88.48} \\
SO-MADD & \textbf{77.65}      & \textbf{87.84}      & \textbf{78.67}       & \textbf{88.69}      & \textbf{79.48}        & 86.89       & \textbf{57.68}      & \textbf{78.60 }    & \textbf{76.78}       & 88.08 \\
\hline
    
FRDM & 75.11 & 81.59 & 74.63 & 84.28 & 74.20 & 84.38 & 55.57 & 76.57 & 74.20 & 86.21 \\
SO-FRDM & \textbf{83.92}      & \textbf{92.37}      & \textbf{77.59}      & \textbf{85.71}      & \textbf{93.04}        & \textbf{89.73}       & \textbf{66.31}      & \textbf{82.83}     & \textbf{78.52}       & \textbf{88.09} \\
\hline
     
RECCE & 79.66 & 89.33 & 77.87 & 87.94 & 79.55 & 91.65 & 53.92 & 70.63 & 67.24 & 83.85 \\
SO-RECCE & \textbf{81.52}      & \textbf{94.77}      & \textbf{85.12}       & \textbf{91.83}      & \textbf{85.85}        & \textbf{95.10}       & \textbf{70.92}      & \textbf{78.41}     & \textbf{83.58}       & \textbf{92.32} \\
\hline
    
CADDM & 74.83 & 84.50 & 73.89 & 80.84 & 72.98 & 83.22 & 56.93 & 74.59 & 70.57 & 80.01 \\
SO-CADDM & \textbf{78.49}      & \textbf{87.96}      & \textbf{74.60}       & \textbf{81.71}      & \textbf{78.18}        & \textbf{86.41}       & \textbf{60.57}      & \textbf{78.25}     & \textbf{75.88}       & \textbf{84.09} \\
\bottomrule
\end{tabular}
}
\label{tab: results_protocol2_retrain}
\end{table*}

\subsubsection{Training Dataset Comparison}
We compare the proposed FFSC dataset with three established face forgery datasets: FF++~\cite{rossler2019faceforensics}, DFDC~\cite{Dolhansky2020deepfake}, and ForgeryNet~\cite{he2021forgerynet} for retraining various face forgery detectors as base models. It is important to note that each dataset may admit a different training strategy. Specifically, FF++ and DFDC treat face forgery detection as a standard binary classification task, while ForgeryNet approaches it as a multi-class classification task. And the FFSC dataset is used to train SO-detectors. To assess cross-dataset generalization, we include two more datasets: DF-1.0~\cite{jiang2020deeperforensics} and Celeb-DF~\cite{li2020celeb}.

Six representative face forgery detectors are selected as base models: Rossler19~\cite{rossler2019faceforensics}, CNND~\cite{wang2019cnngenerated}, MADD~\cite{zhao2021multi}, FRDM~\cite{luo2021generalizing}, RECCE~\cite{Cao_2022_CVPR}, and CADDM~\cite{Dong_2023_CVPR}.
It is important to highlight that some detectors were initially trained with auxiliary tasks, such as image reconstruction~\cite{Cao_2022_CVPR} and manipulation localization~\cite{Dong_2023_CVPR}. To adhere to the original implementation when retraining on FFSC, we incorporate the auxiliary loss (if any) into the objective of the bi-level optimization problem in Eq.~\eqref{eq:auto-weighting}. 
One exception is CADDM, where we omit the multi-scale face swap module because it depends on a reference image that is not accessible during retraining.

The AUC results are shown in Table~\ref{tab: results_differnet_datasets}, from which we find that SO-detectors retrained on FFSC significantly surpass those retrained on the other three datasets. We believe such superior cross-dataset generalization arises because of the label hierarchy of FFSC, which facilitates the SO-detection of face forgery. This approach encourages the detectors to disregard features specific to manipulation methods and instead focus on more transferable features related to face attributes.
Nevertheless, the detectors retrained using FFSC  marginally fall short of those using ForgeryNet~\cite{he2021forgerynet} on the DFDC dataset. This discrepancy may stem from the domain shift attributable to divergent data augmentation techniques and varying photography conditions between FFSC and DFDC. ForgeryNet addresses this gap by employing a broader spectrum of data augmentations, including several that coincide with those used in DFDC. Nonetheless, some augmentations (\eg, highly random brightness adjustment, excessive blurring, and extreme compression) tend to damage major face semantics and are thus not label-preserving.

\begin{table}[!t]
\small
\renewcommand{\arraystretch}{1.2}
\centering 
\caption{AUC results of different face forgery detection methods, in which base detectors are trained on the training set of FFSC. The prefixes ``B-'' and ``M-'' denote the binary and multi-class classification, respectively}
\begin{tabular}{rccccc}
\toprule
Base Model              & FF++ & DFDC & DF-1.0 & Celeb-DF \\
\hline

B-Rossler19 & 87.06 & 74.78 & \textbf{90.39} & 69.12 \\
M-Rossler19 & 85.10 & 72.18 & 86.56 & 75.91  \\
SO-Rossler19 & \textbf{92.99}  & \textbf{77.29} & 89.94  & \textbf{82.23}  \\
\hline  

B-CNND & 80.57 & \textbf{72.99} & 85.98 & \textbf{83.93} \\
M-CNND & 78.14 & 69.84 & 79.31 & 81.32 \\
SO-CNND               & \textbf{88.01}  & 72.37 & \textbf{92.26}  & 83.30  \\
\hline  

B-MADD & 87.63 & 77.02 & 83.96 & 87.46 \\
M-MADD & 86.35 & 76.79 & 83.29 & 84.72 \\
SO-MADD  & \textbf{92.70} & \textbf{78.90}  & \textbf{88.67}  & \textbf{88.61} \\
\hline
   
B-FRDM & 89.07 & 73.58 & 86.46 & 80.39 \\
M-FRDM & 87.65 & 74.18 & 83.96 & 80.87 \\
SO-FRDM & \textbf{93.54}  & \textbf{76.64} & \textbf{86.48}  & \textbf{83.88}  \\
\hline

B-RECCE & 95.56 & 67.20 & 69.73 & 78.51 \\
M-RECCE & 94.55 & 69.23 & 54.81 & 75.94 \\
SO-RECCE & \textbf{97.11} & \textbf{71.78} & \textbf{81.36} & \textbf{80.12} \\
\hline
  
B-CADDM & 83.29 & 65.23 & 87.04 & 81.10 \\
M-CADDM & 85.96 & 69.65 & 84.17 & 79.61 \\
SO-CADDM & \textbf{87.10} & \textbf{71.81} & \textbf{88.08}  & \textbf{83.63}  
\\
\bottomrule
\end{tabular}
\label{tab: results_different_paradigms}
\end{table}

\subsubsection{Training Method Comparison}
To single out the role of our SO-detection method, we compare it against the corresponding base model with the original training strategy on the  FFSC training set.

Table~\ref{tab: results_protocol1_retrain} shows the Protocol-1 results on the additional test set of FFSC (see the descriptions in Sec.~\ref{subsubsec:fm}), which examines the generalization to novel manipulations that alter the same attributes. 
The primary observation is that our SO-detectors significantly improve the efficacy of all base models for almost all face attributes. RECCE~\cite{Cao_2022_CVPR}  performs satisfactorily on its own, without leveraging the label hierarchy in FFSC. This can likely be credited to the integration of a face reconstruction auxiliary task, which promotes extracting semantic face features, thus enhancing generalization across the same face attributes. Frequency-based FRDM~\cite{luo2021generalizing} and distortion-based  CADDM~\cite{Dong_2023_CVPR} both experience noticeable boosts in \texttt{Acc} and \texttt{AUC}, suggesting that the features derived from the SO-detection are either superior to or at least provide a beneficial complement to the signal-level features used in the original methods.

Table~\ref{tab: results_protocol2_retrain} presents the Protocol-2 results, where we train all base models on four out of five face attributes in the training set of FFSC and test them on the remaining face attribute in the test set of FFSC. 
Similarly, we see consistent improvements in performance under the more challenging Protocol-2. Even though the test face attribute is unobserved, our SO-detectors successfully capture the relationships between face attributes by modeling the joint probability distribution, therefore enabling the SO-detectors to generalize across different face attributes.

Moreover, we demonstrate the advantages of our SO-detection method by comparing it to the standard binary and multi-class classification counterparts in Table~\ref{tab: results_different_paradigms}, where we train all base models on the training set of FFSC and test them on FF++~\cite{rossler2019faceforensics}, DFDC~\cite{Dolhansky2020deepfake}, DF-1.0~\cite{jiang2020deeperforensics}, and Celeb-DF~\cite{li2020celeb}. It is evident that the proposed SO-detection method is consistently better across nearly all base models and test datasets. This indicates that exploiting label hierarchy can steer SO-detectors toward learning more generalizable features. Interestingly, multi-class classification tends to hinder detection generalizability, compared to the binary classification baseline. This suggests that multi-class classification appears to promote learning of manipulation-specific cues, which act as ``shortcuts'' for face forgery detection, leading to potential overfitting.

\subsection{Ablation Studies}~\label{subsec: ablations}

\begin{table}[!t]
\renewcommand{\arraystretch}{1.2}
\renewcommand\tabcolsep{4pt}
\centering 
\caption{AUC results of different labeling formulations.
``Global'' denotes the five global face attributes, while ``Local'' stands for the six local face regions. ``Independent'' indicates the equally weighted independent logistic regressions}
\resizebox{\linewidth}{!}{
\begin{tabular}{cclccccc}
\toprule
Global & Local  & Formulation & FF++ & DFDC & DF-1.0 & Celeb-DF & Avg\\
\hline
\cmark &        & Independent & 89.90 & 75.78 & \textbf{91.97}  & 75.74 & 83.35    \\
\cmark &        & Hierarchical & 91.66 & 77.02 & 87.52  & 80.86 & 84.27      \\
\cmark & \cmark & Independent & 89.31 & 71.00 & 91.29  & 71.50 & 80.77    \\
\cmark & \cmark & Hierarchical & \textbf{92.99} & \textbf{77.29} & 89.94  & \textbf{82.23} & \textbf{85.61} 
\\
\bottomrule
\end{tabular}
}
\label{tab: ablations_label_formulation}
\end{table}

\begin{figure}[t]
  \centering
  \includegraphics[width=0.85\linewidth]{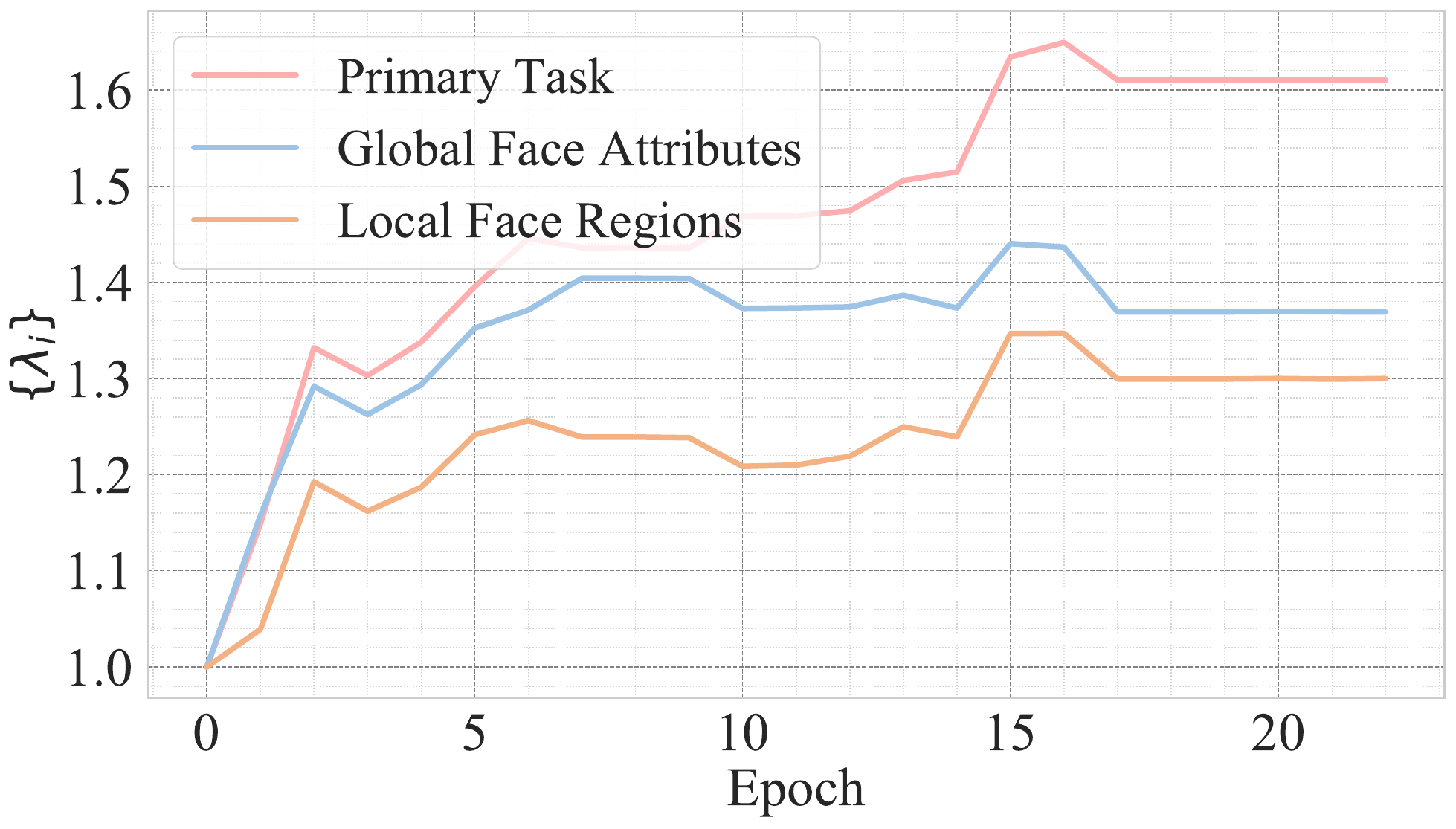}
   \caption{Weight dynamics during bi-level optimization.}
   \label{fig: loss_w}
\end{figure}

\begin{figure*}[t]
  \centering
   \subfloat[Age Manipulation]{\includegraphics[width=0.32\linewidth]{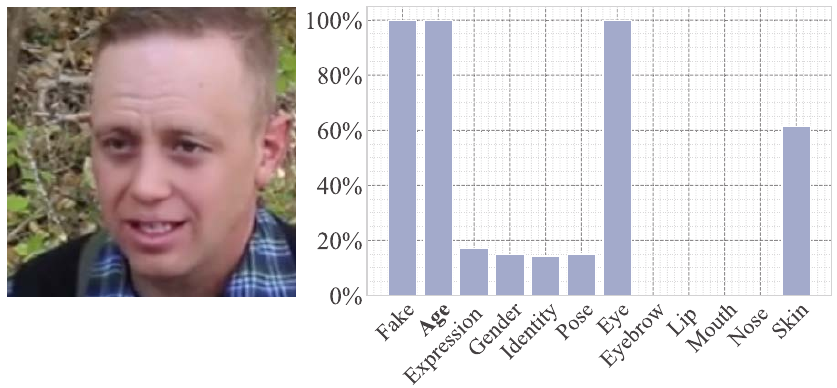}} \hskip.2em
  \subfloat[Expression Manipulation (Smile)]{\includegraphics[width=0.32\linewidth]{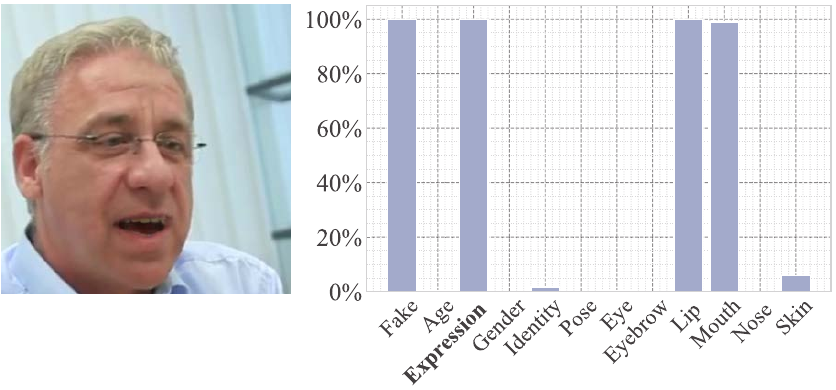}} \hskip.2em
  \subfloat[Expression Manipulation (Surprise)]{\includegraphics[width=0.32\linewidth]{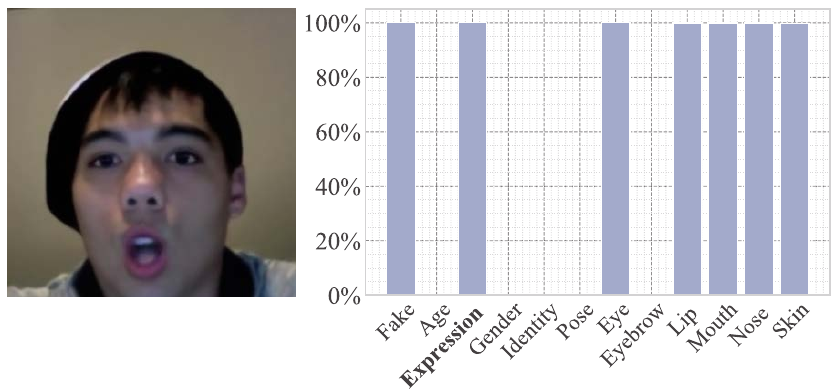}} \hskip.2em
  \\
  \subfloat[Gender Manipulation]{\includegraphics[width=0.32\linewidth]{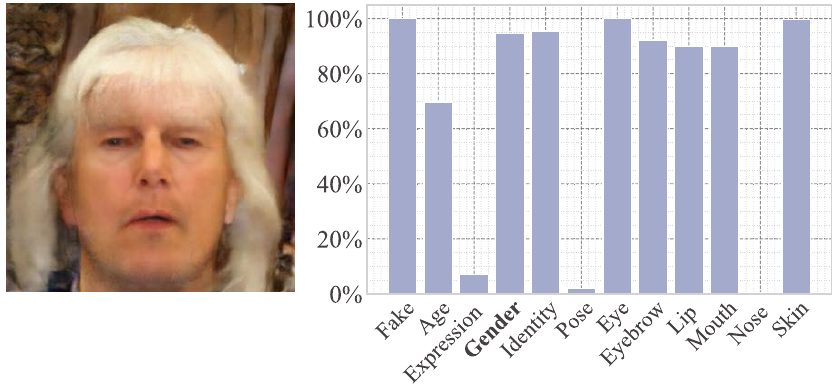}} \hskip.2em
  \subfloat[Identity Manipulation]{\includegraphics[width=0.32\linewidth]{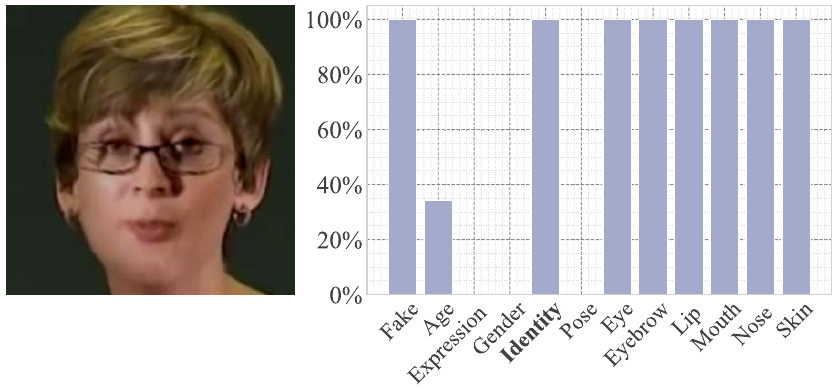}} \hskip.2em
  \subfloat[Pose Manipulation]{\includegraphics[width=0.32\linewidth]{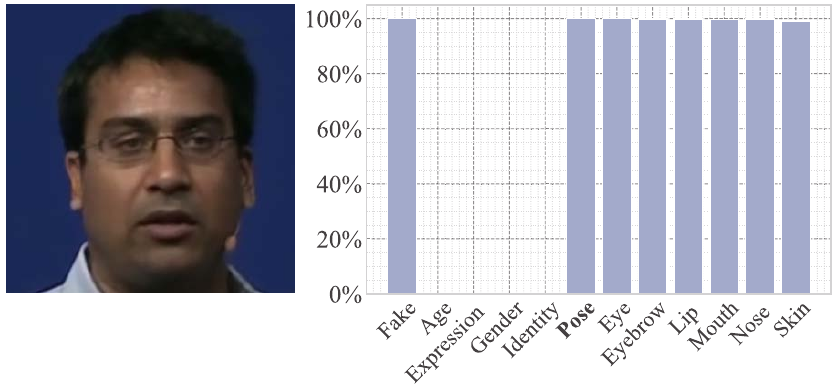}}
  \caption{Illustration of our SO-detector (Rossler19~\cite{rossler2019faceforensics} as the base model) in making predictions at the face attribute and region levels.}
  \label{fig: interpret}
\end{figure*}

We carry out a series of ablation experiments to validate the effectiveness of our SO-detection method using the base model Rossler19~\cite{rossler2019faceforensics} trained on FFSC. We first compare it with the independent logistic regression formulation, where we discard all label relations and weight each task equally. 
Table~\ref{tab: ablations_label_formulation} shows the results, where we find that the proposed label hierarchy significantly enhances detection capabilities. Remarkably, even when we remove the leaf nodes for local face regions, our method still demonstrates substantial efficacy by leveraging the interplay among the five global face attributes.

We next contrast different optimization strategies for leveraging the label hierarchy: 1) minimizing the negative log joint likelihood in Eq.~\eqref{eq: plf},
2) training with the fixed equal weights in Eq.~\eqref{eq: w-plf}, 3) training with the fixed and learned optimal weights (supplied by the bi-level optimization), and 4) training with dynamic weighting average~\cite{liu2019end} without prioritizing the primary task. 
As presented in Table~\ref{tab: ablations_loss_weighting}, direct optimization of the joint likelihood of all observed labels results in a performance comparable to fixed equal weighting, with neither method giving prominence to the primary task. The application of dynamic weighting average appears ineffective in learning to prioritize the primary task, yielding only a marginal improvement over fixed equal weighting. Furthermore, our findings indicate that employing fixed optimal weights determined through bi-level optimization falls short of matching the performance of the default bi-level optimization. This underscores the significance of dynamic weighting adjustment in promoting the primary task during optimization (see Fig.~\ref{fig: loss_w}).

\begin{table}[!t]
\renewcommand{\arraystretch}{1.2}
\centering 
\caption{AUC results of different optimization strategies}
\resizebox{\linewidth}{!}{
\begin{tabular}{lccccc}
\toprule
Optimization strategy  & FF++ & DFDC & DF-1.0 & Celeb-DF & Avg \\
\hline
Joint likelihood  & 88.86 & 75.28 & 87.48  & 81.67 &   83.32  \\
Fixed equal weights & 86.35 & 76.40 & 88.32  & 81.27 & 83.09         \\
Fixed learned weights & 91.86 & 76.94 & 89.73  & 81.01 &  84.88   \\
Dynamic weighting average & 90.20 & 75.70 & \textbf{90.31}  & 79.11 & 83.83      \\
\hline
Bi-level optimization (Ours) & \textbf{92.99} & \textbf{77.29} & 89.94  & \textbf{82.23} & \textbf{85.61}        \\
\bottomrule
\end{tabular}
}
\label{tab: ablations_loss_weighting}
\end{table}

\begin{figure}[t]
  \centering
  \subfloat[Gender manipulation]{\includegraphics[width=0.9\linewidth]{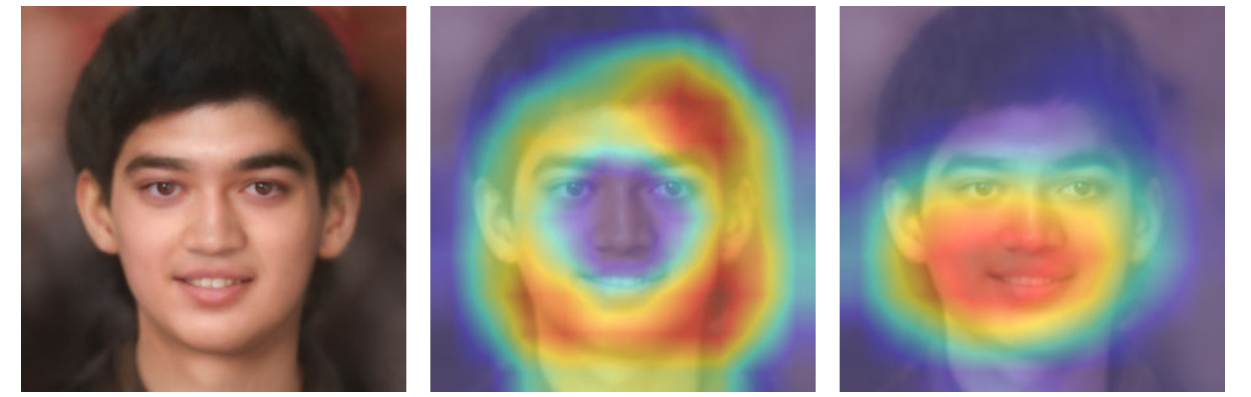}}\\
  \vspace{-.2cm}
  \subfloat[Identity manipulation]{\includegraphics[width=0.9\linewidth]{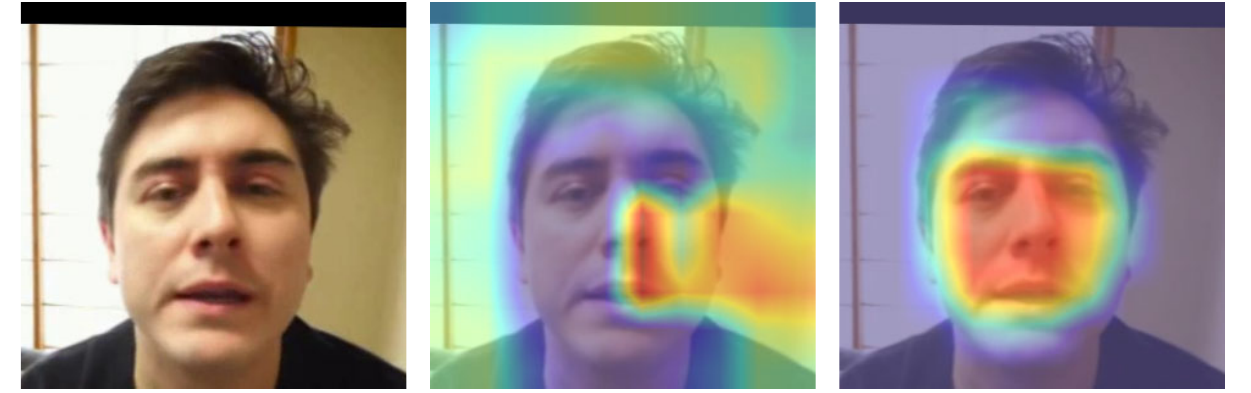}}\\
  \vspace{-.2cm}
  \subfloat[Pose manipulation]{\includegraphics[width=0.9\linewidth]{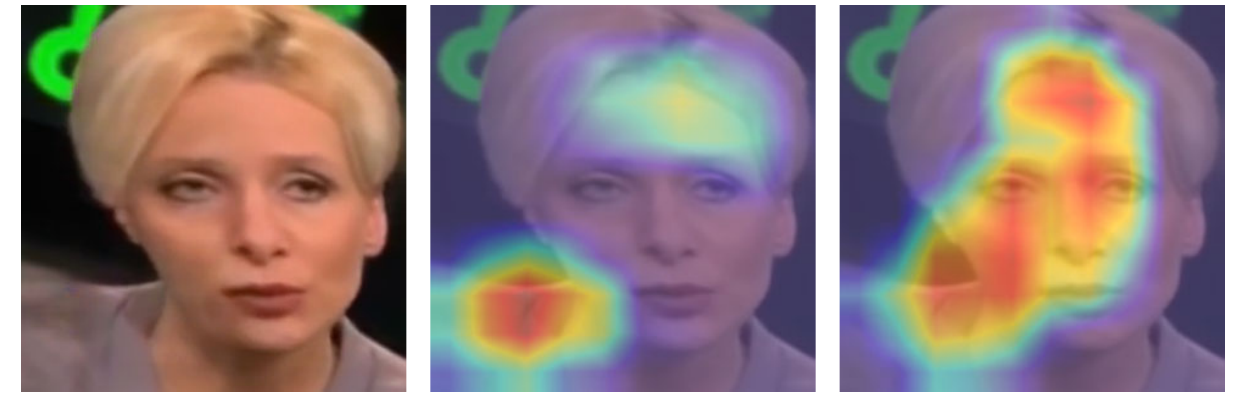}}
  \caption{Visualization of Grad-CAM activation maps. Each subfigure shows the forged face (left), base Rossler19~\cite{rossler2019faceforensics} heatmap (middle), and SO-Rossler19 heatmap (right).}
  \label{fig: grad_cam}
\end{figure}

\subsection{Further Analysis}
Benefiting from the proposed SO-detection method, we can compute the marginal probability of each face attribute and region being manipulated, which endows the SO-detectors with some degree of interpretability. For example, in Fig.~\ref{fig: interpret}(b), our method discerns the face as fake, and meanwhile indicates a likely alteration of expression, pinpointing the mouth and lip as manipulated regions.

Additionally, we use Grad-CAM~\cite{selvaraju2017grad} to visualize the activation maps of the base Rossler19~\cite{rossler2019faceforensics} and SO-Rossler19 in 
Fig.~\ref{fig: grad_cam}. The results demonstrate that the SO-detector consistently targets meaningful semantic face regions, whereas the baseline detector frequently depends on ``cheap shortcuts,'' such as background artifacts for detection.

\section{Conclusion and Discussion} \label{sec: conclusion}
In this work, we have given face forgery a new definition from the perspective of face semantics. Based on this definition, we constructed a new face forgery image dataset, FFSC, and proposed a SO-detection method. Extensive experiments have validated the promise of the proposed FFSC dataset as training and test sets, and the superiority of our method in improving the generalizability of face forgery detectors.

It is important to note that the current definition of face forgery is imperfect. One counterexample is that reducing the frame rate in a speech video of House Speaker Nancy Pelosi makes her sound sluggish and slurred. Despite the obvious falseness of the video, it would still be classified as real under our definition as it does not involve any modifications of semantic face attributes. To address this issue, a possible solution could be adding a \texttt{psychological\_response} node at the global face attribute level. Additionally, it is natural to expand the label hierarchy in Fig.~\ref{fig_DAG}(a) by adding other attribute nodes (\eg, \texttt{race} and \texttt{attractiveness}). 
Additionally, the FFSC dataset could integrate other modalities, such as text, to enable multimodal face forgery detection. In such cases, multimodal large-language models could be used to generate semantically localized descriptions of face manipulations~\cite{shao2023dgm4}.

\begin{figure}[t]
  \vspace{-.3cm}
  \centering
  \subfloat[]{\includegraphics[width=0.5\linewidth]{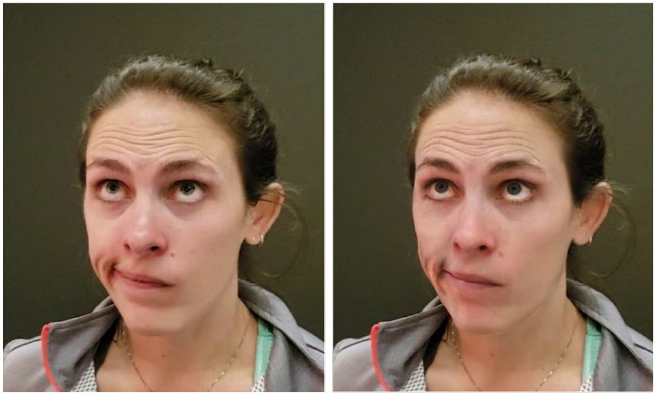}} 
  \subfloat[]{\includegraphics[width=0.5\linewidth]{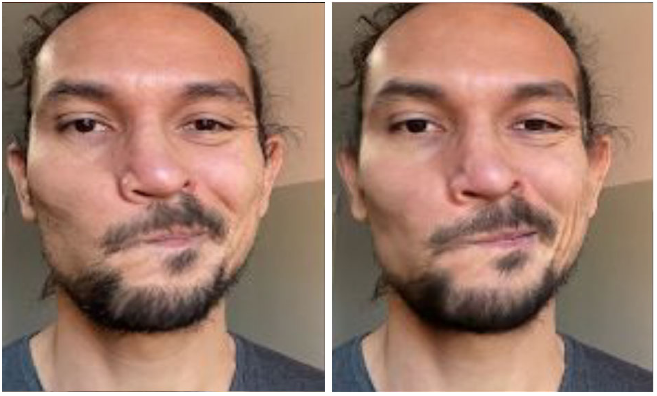}} 
  \caption{Face rendering results by the method in \cite{chen2021high}. Within each subfigure, the real and rendered faces are displayed on the left and right, respectively.}
  \label{fig: face_render}
\end{figure}

\begin{figure}[t]
  \centering
  \includegraphics[width=\linewidth]{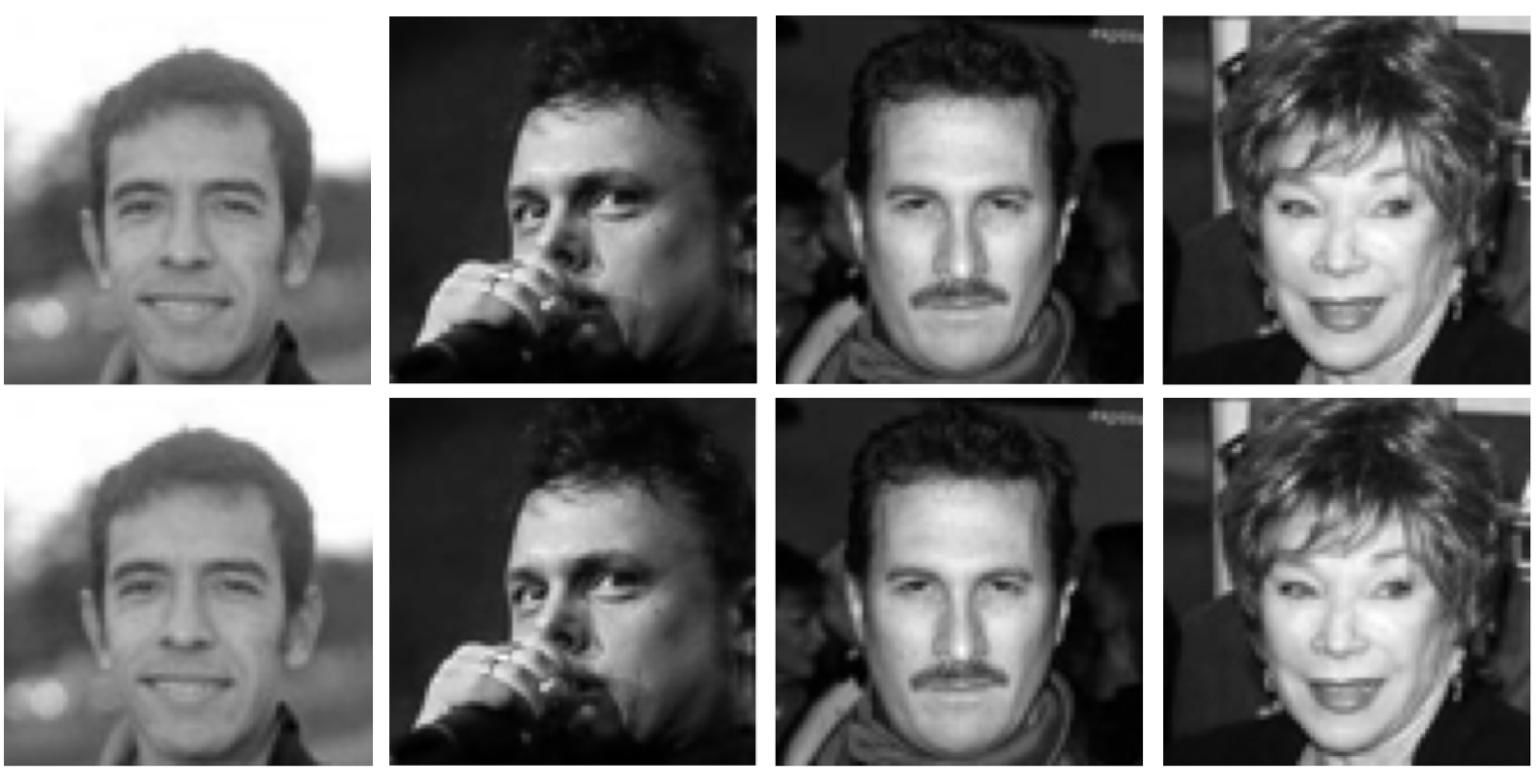}
  \caption{
  Memorization results~\cite{kadkhodaie2023generalization} of a diffusion model trained on a small-sized CelebA dataset~\cite{liu2015faceattributes}. The top row depicts training face images downsampled to $80\times 80$, and the bottom row displays face images generated by the diffusion model, which closely resemble those from the training set.  
  }
  \label{fig: memorization}

\end{figure}

Our current definition of face forgery is limited to face manipulation methods.  It would be beneficial to expand this definition to include face creation methods, such as those utilizing GANs~\cite{karras2020analyzing}, diffusion models~\cite{song2019generative, rombach2022high, bhattacharyya2024diffusion} and computer graphics~\cite{chandran2021rendering}. However, this expansion presents significant challenges. First, technology has advanced to the point where we can render an existing face presented in a real physical scene with such realism that it becomes indistinguishable from an actual photograph (see Fig.~\ref{fig: face_render}).  Second, there is evidence that generative models can memorize training data~\cite{kadkhodaie2023generalization}, and techniques exist to retrieve these memorized images (see Fig.~\ref{fig: memorization}). In both cases, determining the authenticity of a face image requires heightened awareness or insights to perceive its true nature. Taking a step back, a more pragmatic extension of our definition is to include digital manipulation methods for images of natural scenes that do not necessarily contain faces.

\section*{Acknowledgements}
The authors would like to thank Mr. Weiran Zhao for his kind help in data collection. This work was supported in part by the Hong Kong RGC General Research Fund (11220224), the CityU Strategic Research Grants (7005848 and 7005983), and an Industry Gift Fund (9229179).

\bibliographystyle{IEEEtran}
\bibliography{FFSC}

\begin{IEEEbiography}[
{\includegraphics[width=1.5in,height=1.3in,clip,keepaspectratio]{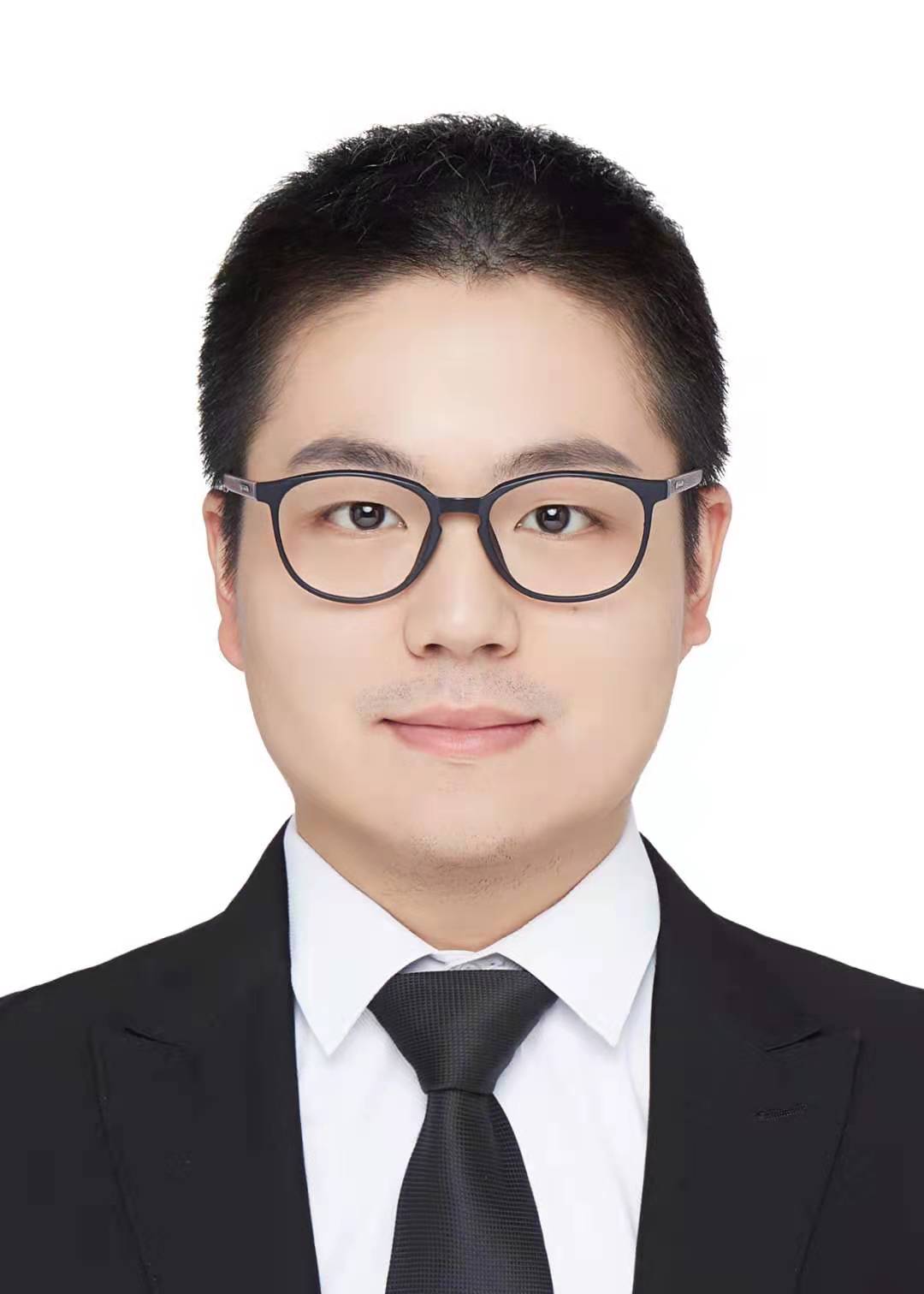}}]{Mian Zou}
received the B.E. degree from Hefei University of Technology, Hefei, China, in 2018, and the M. Eng. degree from the University of Shanghai for Science and Technology, Shanghai, China, in 2021. He is currently pursuing a Ph.D degree with the Department of Computer Science at the City University of Hong Kong, Kowloon, Hong Kong. His research interests include multimedia forensics and computer vision.
\end{IEEEbiography}
\vspace{-4em}

\begin{IEEEbiography}[
{\includegraphics[width=1.5in,height=1.3in,clip,keepaspectratio]{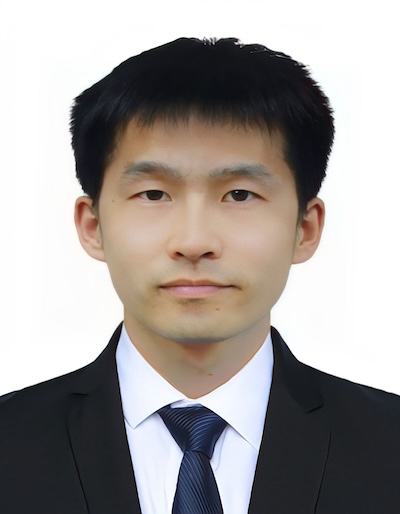}}]{Baosheng Yu}
received the B.E. degree from the University of Science and Technology of China, Hefei, China, in 2014, and the Ph.D. degree from the University of Sydney, Camperdown, NSW, Australia, in 2019. He is currently an Assistant Professor with the Lee Kong Chian School of Medicine at Nanyang Technological University, Singapore. He has authored or coauthored more than 40 publications on top-tier international conferences and journals, including CVPR, ICCV, ECCV, and IEEE Transactions on Pattern Analysis and Machine Intelligence. His research interests include
computer vision and machine learning.
\end{IEEEbiography}
\vspace{-4em}

\begin{IEEEbiography}[
{\includegraphics[width=1.5in,height=1.3in,clip,keepaspectratio]{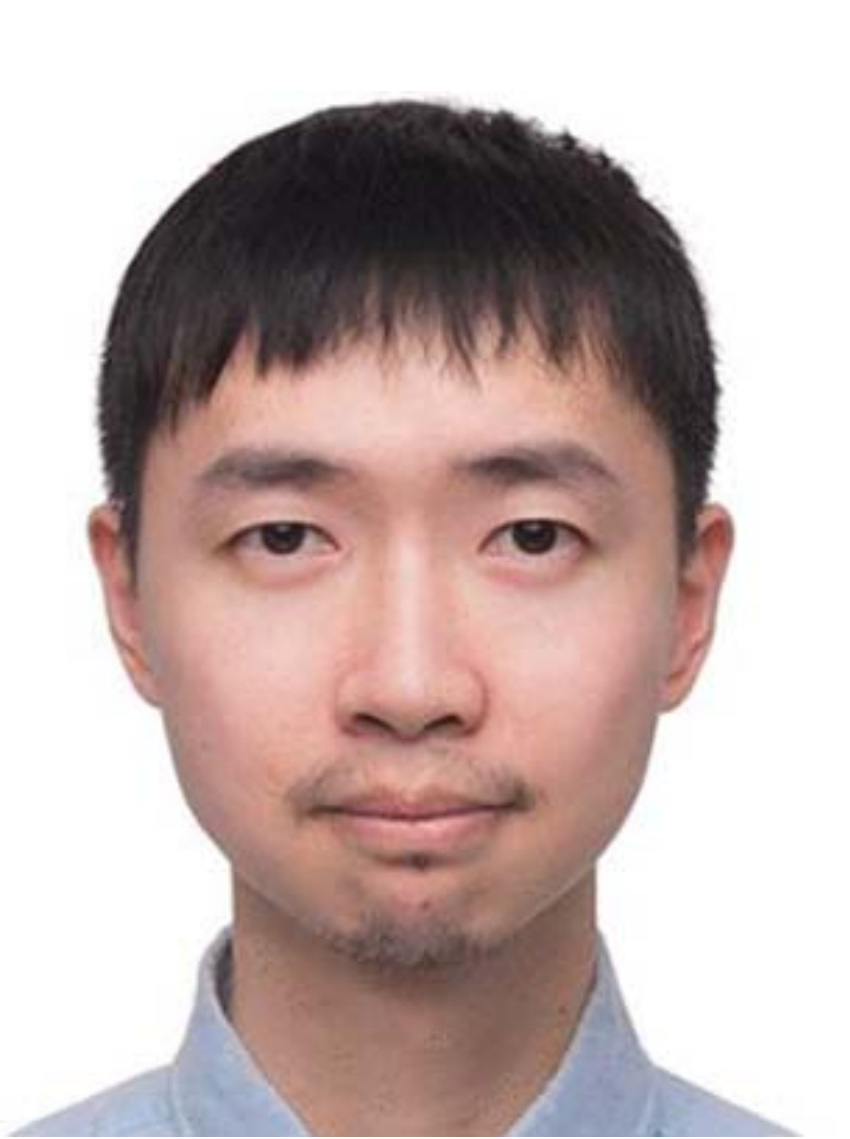}}]{Yibing Zhan} (Member, IEEE) received the B.E. degree and Ph.D. from the University of Science and Technology of China in 2012 and 2018, respectively. From 2018 to 2020, Yibing Zhan worked as an associate researcher at the School of Computer Science, Hangzhou Dianzi University. He is currently an Algorithm Scientist with the JD Explore Academy. His research interests include scene graph generation, foundation models, and graph neural networks. He has authored or coauthored many scientific papers in top conferences and journals, such as NeurIPS, CVPR, ACM MM, ICCV, and IEEE Transactions on Multimedia.
\end{IEEEbiography}
\vspace{-4em}

\begin{IEEEbiography}[
{\includegraphics[width=1.5in,height=1.3in,clip,keepaspectratio]{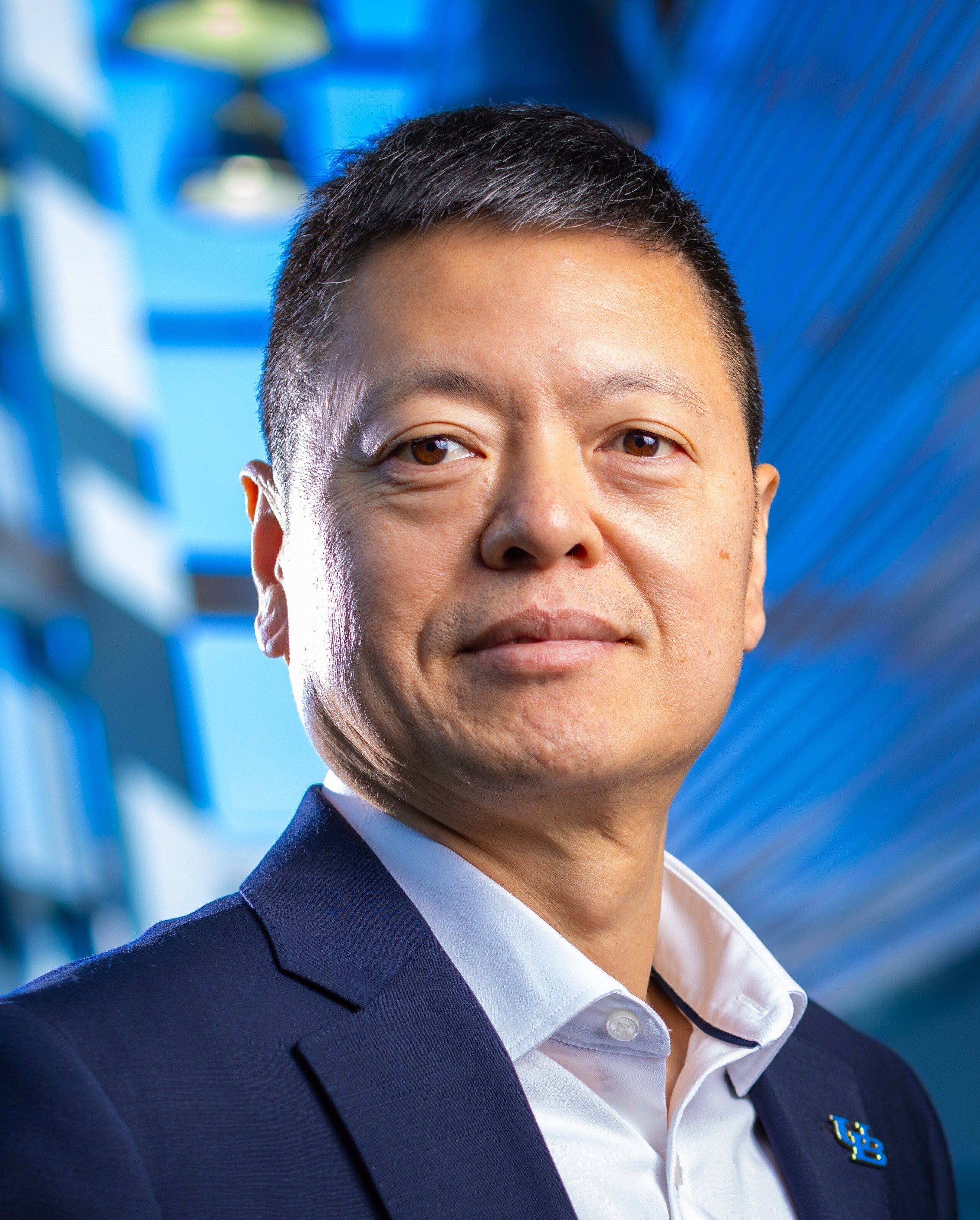}}]{Siwe Lyu} (Fellow, IEEE) received the B.S. and M.S. degrees in computer science and information science from Peking University, Beijing, China, in 1997 and 2000 respectively, and the Ph.D. degree in computer science from Dartmouth College, Hanover,
NH, USA, in 2005. He is currently a SUNY Empire Innovation Professor with the Department of Computer Science and Engineering, University at
Buffalo, State University of New York at Buffalo, Buffalo, NY, USA. His research interests include digital media forensics, computer vision, and machine learning. He is a Fellow of IEEE, IAPR and AAIA, and a Distinguished Member of ACM.
\end{IEEEbiography}
\vspace{-4em}

\begin{IEEEbiography}[
{\includegraphics[width=1.5in,height=1.3in,clip,keepaspectratio]{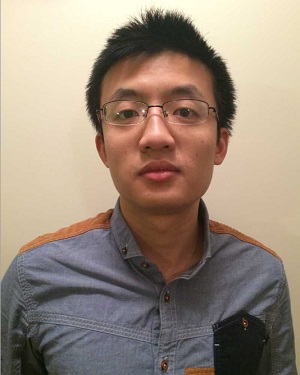}}]{Kede Ma} (Senior Member, IEEE) received the B.E. degree from the University of Science and Technology of China, Hefei, China in 2012, and the M.S. and Ph.D. degrees in electrical and computer engineering from the University of Waterloo, Waterloo, ON, Canada in 2014 and 2017, respectively. He was a Research Associate with the Howard Hughes Medical Institute and New York University, New York, NY, USA in 2018. He is currently an Assistant Professor with the Department of Computer Science at the City University of Hong Kong. His research interests include perceptual image processing, computational vision, computational photography, multimedia forensics and security, and machine learning for multimedia signals.
He currently serves on the Editorial Boards of IEEE Transactions on Image Processing and IEEE Transactions on Information Forensics and Security.
\end{IEEEbiography}

\end{document}